\documentclass{llncs}

\usepackage{hyperref}
\usepackage{comment}
\usepackage{dsfont}
\usepackage{stmaryrd}
\usepackage{graphicx}
\usepackage{wrapfig}

\newcommand{\set}[1]{\{ #1 \}}

\begin{document}

\title{Learning Causality: \\
	Synthesis of Large-Scale Causal Networks \\\
	from High-Dimensional Time Series Data}
\author
{
	Mark-Oliver Stehr\inst{1} \and 
	Peter Avar\inst{2} \and
	Andrew R. Korte\inst{2} \and 
	Lida Parvin\inst{2} \and \\
	Ziad J. Sahab\inst{2} \and 
	Deborah I. Bunin\inst{1} \and 
	Merrill Knapp\inst{1} \and 
	Denise Nishita\inst{1} \and \\
	Andrew Poggio\inst{1} \and 
	Carolyn L. Talcott\inst{1} \and 
	Brian M. Davis\inst{3} \and 
	Christine A. Morton\inst{3} \and 
	Christopher J. Sevinsky\inst{3} \and
	Maria I. Zavodszky\inst{3} \and
	Akos Vertes\inst{2} 
}

\institute
{
	SRI International, Menlo Park, CA 94025\and
	Dept. of Chemistry, George Washington Univ., Washington, DC 20052\and 
	GE Global Research, Niskayuna, NY 12309
}

\date{March 2019}

\maketitle

\begin{abstract}
	There is an abundance of complex dynamic systems that are critical to our daily lives and our society but that are hardly
	understood, and even with today's possibilities to sense and collect large amounts of experimental data, they
	are so complex and continuously evolving that it is unlikely that their dynamics will ever be understood in full detail.
	Nevertheless, through computational tools we can try to make the best possible use of the current technologies and
	available data. We believe that the most useful models will have to take into account the imbalance between system complexity and available data in the context of limited knowledge or multiple hypotheses. The complex system of biological cells is a prime example of such a system that is studied in systems biology and has motivated the methods presented in this paper. They were developed as part of the DARPA \textit{Rapid Threat Assessment} (RTA) program, which is concerned with understanding of the \textit{mechanism of action} (MoA) of toxins or drugs affecting human cells. Using a combination of Gaussian processes and abstract network modeling, we present three fundamentally different machine-learning-based approaches to learn causal relations and synthesize causal networks from high-dimensional time series data. While other types of data are available and have been analyzed and integrated in our RTA work, we focus on transcriptomics (that is gene expression) data obtained from high-throughput microarray experiments in this paper to illustrate capabilities and limitations of our algorithms. Our algorithms make different but overall relatively few biological assumptions, so that they are applicable to other types of biological data and potentially even to other complex systems that exhibit high dimensionality but are not of biological nature.
\end{abstract}

\section{Introduction and Motivation}

Systems biology strives at an understanding of all the interactions between components of a biological system over time. The excellent survey \cite{Zak09} describes some of the challenges and opportunities in the context of human cells, using the innate immune system as a prime example. It is argued that systems biology has become a highly interdisciplinary endeavor that involves an iterative cycle which is driven by the interplay between emerging biological problems and the development of new technologies and computational tools. 
One particular approach to abstract from the complexity of biological systems arises from network science \cite{Gosak18}. In this context, a wide range of graph-based formalisms, ranging from broad
classes such as scale-free graphs \cite{Albert05} to very concrete formalisms such as variations of Petri nets \cite{Chaouiya07} have been explored. While the comprehensive modeling of, say a human cell as a network, and hence its response to a wide range of conditions, in other words, addressing the system identification problem, is an important long-long term objective of system biology, we are concerned with a more modest goal in this paper, namely the modeling of key processes in the human cell under very specific conditions and at a level of abstraction that is consistent with the large but in many ways still limited amount of data that is available with today's high-throughput measurement technologies. As part of an iterative cycle such as the one discussed in \cite{Zak09} these models can lead to hypothesis generation, inform new experiments, and can be subsequently confirmed and possibly refined or rejected based on new results.

More specifically, we present some of the algorithms developed in the DARPA \textit{Rapid Threat Assessment} (RTA) program, where we have been exploring data analysis, machine-learning, and logic-based techniques to support biologists in understanding the so-called \textit{mechanism of action} (MoA) that is triggered when an (unknown) drug or toxic substance hits a human cell. From relatively short windows of time after the event in question (e.g. 48 hours) our algorithms generate graphs representing potential causality between compounds. In spite of the use of perturbations, it should be noted that this abstract notion of causality is based on obervational data with its known limitations (e.g. confounding effects), and might be better called causality modulo observational equivalence. This is in contrast to, for example, knock-out studies for individual genes, which however due to higher cost cannot compete with the sheer data volume and coverage typical for observational studies. The basis for our algorithms are time series of typically high-dimensional data, e.g. transcriptomics (gene expression), proteomics, and metabolomics data. As exemplified in \cite{Vertes18}, we have also developed algorithms for anomaly detection that highlight certain nodes in such graphs as potentially impacted and allow the biologist to narrow down the mechanism of action. The algorithms developed use a variety of models including Gaussian processes (on non-linear time scales) and other linear and non-linear models, ranging from principal component analysis and various types of clustering to a broad range of neural network models. For more details about RTA and some initial results we refer to \cite{Vertes18}.

More recent algorithms that we developed include anomaly detection using different types of autoencoders, causality detection and network graph synthesis based on convolutional autoencoders, deep and wide neural networks predicting temporal evolution and their visualization as graphs, generative adversarial networks for synthetic modeling and detection of typical vs.\ unusual behavior, and Siamese (twin) neural networks for probabilistic causality detection. 

In this paper, the primary focus is on Siamese neural networks for probabilistic causality detection
and their training and validation using  a synthetic dynamic gene expression model taking advantage of our original Gaussian process model. They are the basis for two models that support the synthesis of probabilistic causal networks, meaning that causality can be probabilistically quantified. The price to pay for probabilistic results lies in some underlying modeling assumptions about the structure and dynamics of these biological networks. Our other algorithms for network synthesis offer a non-probabilistic notion of potential causality that comes with its own advantages and disadvantages. Correlation and cross-correlation graphs such as those in \cite{Vertes18} fall into this heuristic category, but also our autoencoder-based network synthesis and network synthesis based  on predictive deep and wide neural networks. 
A discussion of these heuristic approaches is included in this paper as well.

While our work is directly motivated by the systems biology domain, we conjecture that our methods are applicable in other domains where high-dimensional observational time series data is available and network modeling is essential. For concreteness, however, we continue to use the biological domain in our explanations, specifically focussing on transcriptomics (gene expression) data generated by high-throughput microarray experiments, as this is the richest data source available in the RTA project and as it allows us to relate our results to those of other projects to asses biological plausibility.

As we will see in the following section on related work, there are important distinctions between the problem of interest, namely identifying the mechanism of action (MoA), and other related problems such as the reconstruction of gene regulatory networks. First of all, our goal is to use observational data from a single experiment in the best possible way to narrow down the MoA to a small set of candidates, which may be used to inform further experimentation. Reconstruction of the entire network from such limited experimental conditions is clearly not feasible, because only a small number of network interactions may have been excercised and hence observed under those conditions. Since we are only concerned with networks in this paper, we also adopt an abstract and greatly simplified notion of MoA as a network of potential causality. We assume that for a given experiment such an MoA (process) network should be consistent with and hence a subnetwork (another simplifying assumption) of the unknown biological (system) network, for example the gene regulatory network if we just focus on transcriptomics. Note how we distinguish the system-level view --- a system-level biological model would capture all possible behavior at a certain level of abstraction --- from the process-level view, which captures the (abstract) biological behavior under specific experimental (including initial) conditions and is the focus of
our analysis. Another noteworthy feature of our approach is that it minimizes the amount of prior biological knowledge
used in the analysis, for example unlike \cite{Ramsey08} we do not make assumptions about which genes act as transcription factors  and we do not use drug signature databases such as \cite{Lamb06} or other databases curated from the literature \cite{Han15}. This is motivated in the context of the RTA project by the initial lack of knowledge about new threats/mechanisms that are unknown or just being discovered. Of course, this does not preclude the combination of our methods with prior knowledge in other applications.

\section{Related Work}

The Connectivity Map \cite{Subramanian17} is a database of differential gene expression signatures, that is transcriptional responses to disease, genetic perturbation (knockdown or overexpression of a gene) or treatment with a small molecule. In its current version the library contains more than 1 million signatures resulting from perturbations of multiple human cell types. Connectivity between signatures refers to their similarity or dissimilarity. The latest version of the Connectivity Map is based on the L1000 assay, which is a Luminex bead-based method that measures the mRNA expression of 978 so-called landmark genes, and uses a machine-learning-based inference algorithm to infer the expression of 11350 additional genes. An earlier version of the Connectivity Map \cite{Lamb06} was based on the direct measurement of the full transcriptome using microarrays, but was limited to $6100$ combinations of perturbations and cell types. Our RTA project also uses the microarray-based method to measure the full transcriptome. Different from the above, we apply it to collect time series data, but due to the higher cost focus only on a small number of perturbation experiments (seven experiments have been conducted so far with six different drugs). We also collect corresponding time
series of proteomics and metabolomics data for a more comprehensive picture.

The DREAM project (Dialogue on Reverse Engineering Assessment and Methods) conducted an assessment of over thirty gene regulatory network inference methods on E. coli, S. aureus, S. cerevisiae, and in silico microarray data \cite{Marbach12,Prill10}. The evaluation indicates that there is no single best method for all data sets. As a possible solution, an integration of this community of network inference experts is proposed, leading to an integrated community network for each data set. The in-silicio data set \cite{Prill10} is particularly noteworthy, since it is based on a well-defined synthetic network combining subnetworks from E. coli and S. cerevisiae with a continuous differential equation model and Gaussian noise. Note, however, that none of the networks are representative of the large size and interaction complexity of the human gene regularity network, which is furthermore unknown to a large extent. Moreover, each data set contains knock-out and know-down experiments, which facilitates the detection of dependencies. As an example, the inferred community network for E. Coli contains 1688 edges. About $50\%$ of the edges were known and previously experimentally validated interactions, while the remaining $50\%$ were conjectured to be false positives or newly discovered true interactions (which were confirmed by additional experiments). The algorithms developed in several DREAM challenges quite clearly demonstrate the challenges of the system identification problem even under very favorable conditions, that is small overall network size, relatively low dynamic complexity due to non-human cells, and a reasonably good understanding of ground truth. While we use human cells, we focus on the identification of the process (MoA) under very specific conditions, and the synthetic dynamic model we use is very generic and kept as simple as possible to reduce the assumptions and number of parameters that need to be stipulated.

Our probabilistic approach to causal network synthesis is quite different from the traditional approach of \cite{Spirtes93,Pearl00} and hence the large body of research conducted in the context of Bayesian networks and their many variations. While these approaches, are primarily based on exploiting certain asymmetries in the structure of conditional independence relations (involving at least three variables) that partly reveal the direction of causality, our approach is focused on directly detecting binary causality based on learned patterns occurring in a high-dimensional time series. Our focus on relations between only two variables enables better scalability and more directly focuses on the relation of interest and its physical nature. Graphs with a million causal edges are not unusual in our targeted applications, even if the user might only focus on a small subset at a time. While the traditional approach is also applicable to time series (see e.g. \cite{Chu08}), they need to be sufficiently long to reliably perform the independence tests, in other words, they ideally need to exercise the system in a large number of conditions, which is not the case in our targeted applications. While
in Bayesian network inference the consistency of edge orientation is a primary concern, we intentionally avoid imposing consistency constraints in the algorithms presented in this paper. In a separate paper \cite{palo}, we show how domain-dependent consistency requirements can be implemented in an approximate and scalable fashion on top of the probabilistic causal graph synthesis presented here.

The particular artificial neural networks we use for causality detection in our primary method fall into the general class of Siamese neural networks, which are composed using two subnetworks with shared weights, each responsible for processing one of the inputs. While originally proposed for signature verification (a type of similarity detection) in
\cite{Bromley93,Bromley94}, they have recently attracted renewed interest \cite{Koch15,Shaham15,Berlemont18,ShahamL18,Manocha18,Zhang18,Droghini18,Li19}, with many applications
ranging from image recognition to audio search. The key abstract feature of Siamese networks that is important for
us is their capability to directly encode structural properties of a logical relation (in our case \textit{undirected} causality) and hence to enable efficient learning tailored to the domain. Although we use relatively low-complexity 1D convolutional Siamese networks suitable for time series processing, we had to resort to curriculum training to obtain a stable learning algorithm. While conventional Siamese networks are suitable for similarity relations that exhibit a structural symmetry,
we also use a variation of Siamese networks to detect \textit{directed} causality which directly encode the
structural antisymmetry of this relation (when viewed as a function) in terms of the network structure.

\section{Probabilistic Causal Network Synthesis}

In this section, we informally summarize our probabilistic approach to network synthesis. First, we briefly review our use of Gaussian processes, which are at the core of our broader RTA workflow and also the foundation for our synthetic dynamic model. This model will be used to train and validate our undirected and directed causality detectors (both different types of Siamese neural networks in the broad sense). Undirected and directed causal network synthesis will then be performed based on these detectors. For illustration purposes, we use primarily results from our RTA workflow for the $Unk5$ experiment \cite{Vertes18}, which was used for a DARPA challenge and later revealed to be a statin (specifically atorvastatin, a drug affecting cholesterol biosynthesis). Finally, we conclude this section with a biological plausibility check that is based on entirely independent experimental data from two other large projects.

\subsection{Gaussian Processes}

A Gaussian process \cite{gaussianprocesses} represents a \textit{probability distribution} over a \textit{parameterized class of continuous functions}. The projection at each time point is Gaussian, i.e., normally distributed, with a \textit{mean and variance that is explicitly represented} in the Gaussian process model. A time series of observations can then be obtained by sampling at a finite number of arbitrary time points.

\begin{figure*}[!htb]
	\includegraphics[width=1.0\linewidth]{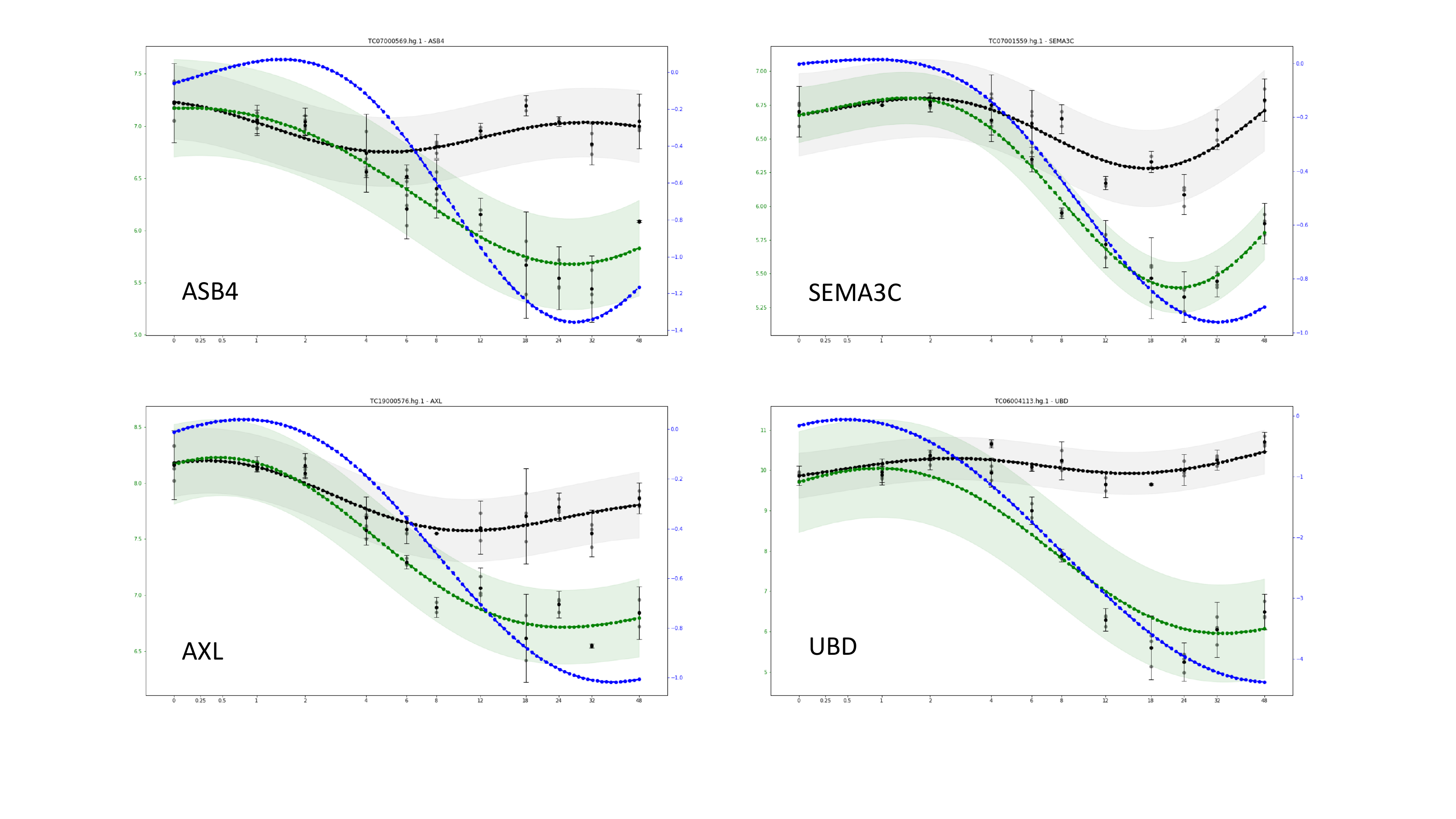}
	\caption{Sample Gaussian processes for control (black) and treated (green) with 2 SD bands and the log-ratio shown in blue (right scale). }
	\vspace{5pt}
	\label{fig-gp}
\end{figure*}

Gaussian process modeling allows us to address several challenges when faced with high-dimensional and very noisy biological time series data. First, observations are only available at a small number of time points with irregular spacing between them. Also the number of observations at each time point may vary. By exploiting the continuity of the underlying functions, Gaussian processes allow us to benefit from the local structure to estimate the mean and to provide an explicit estimate of variance at each time point. The second challenge arises in the integration of multiple data sources, in our case we have transcriptomic, proteomic, and metabolomic data, each with different sampling time points, and different amounts of measurement noise. Again, the Gaussian processes allow us to integrate all data into a uniform continuous time scale and allow us to build specialized models for each data source to reflect the \textit{biological variation} (over time) and the \textit{measurement error} (at each time point).

Specifically, we use \textit{GPflow} \cite{gpflow}, a library based on TensorFlow, to estimate Gaussian processes separately for each type of data, for each compound (transcript, protein, or metabolite), and for treated and control conditions. All observed quantities are uniformly represented in log2 scale. Furthermore, we use a log-based time scale to reflect the approximate spacing of observations with more observations closer to the beginning of the experiment (defined as Time 0). The class of functions is generated by a kernel that combines squared exponentials (reflecting biological variation) in the log-based time scale with additive white noise (to capture measurement error). The squared exponential time scale parameter is fixed at $2$ to reflect the density of available observations. The squared exponential variance is fixed for each type of data by using an average of the optimal values over all compounds. The white noise variance is computed by Maximum Likelihood Estimation (MLE) for each specific compound. 

The output of the Gaussian process workflow is a uniform time series for each compound with $101$ time points ($100$ time points without the initial state) for control and treated sample means and the corresponding standard deviations (SDs). Formally, the learned Gaussian process model is continuous.  The $101$ time points sample the time dimension at sufficiently many discrete points to preserve the information. The \textit{log-ratio time series} for each compound, which is the primary basis for subsequent stages of our workflow, is then computed by point-wise difference between treated and control.

In Fig.~\ref{fig-gp} we show Gaussian processes for four selected genes and their log-ratio in the $Unk5$ experiment. Do the similar shapes indicate the presence of a causal relationship and if yes in which direction ? The answer is not obvious due to large amount of noise, and clearly has to be probabilistic in nature. These are the kind of questions we are trying to address in the following subsections.

The confidence in the change (as measured by the log-ratio) in response to the perturbation is scored by computing the area between the SD bands for control and treated samples. We perform the computation for 1 and 2 SDs, the former to account for the higher degree of noise in the proteomics and metabolomics data. This provides a \textit{ranking of the measured compounds} that generalizes the biologist criteria that for a significant change the SDs of the control and treated samples should not overlap. This ranking can be used as a first filter to focus attention on compounds in which we have some confidence in the measured change.

\subsection{Synthetic Dynamic Model}

To train our probabilistic causality detector, we use a simple synthetic model determined by a small number of parameters.
The model needs to provide positive and negative examples of causally related genes in the context of a given experiment. To this end we assume that the previously estimated Gaussian processes for each gene represents a distribution of biological plausible gene expression profiles that are consistent with the observed data. Our first step is to generate a set of noisy pairs of control and treated time series, that is a set of pairs $(c_g,u_g)$, where $c_g$ and $u_g$ are sampled from the Gaussian process model for control and treated (e.g. with an unknown drug or toxin in the context of our RTA challenges) conditions of a random gene $g$. For each pair, we compute the difference $r_g = u_g - c_g$, giving rise to a set of log-ratio time series samples $r_g$ consistent with the observed data. In another step, each sampled ratio time series will be normalized, so that that all have mean $0$ and variance $1$.
We use $\bar{r}_g$ to denote a sample normalized series.

Our first goal is to construct synthetic gene expression time series for causally related pairs  (positive examples). 
To obtain a uniform size time series representation for the synthetic genes, we fix a window size $w$ smaller than the size of the full time series (e.g. $80$ of out $101$).\footnote{The window size is a fixed hyperparameter that reflects a tradeoff between the number of time lags we can model and the data available for causality detection.}  To simulate a possible time lag associated with causality, we sample $\bar{r'}_g$ and $\bar{r}_g$ from the previous normalized log-ratio time series set, and compute the \textit{set of lagged pairs} $(\bar{l}'_g,\bar{r}_g)$ where $\bar{l}'_g$ is obtained from $\bar{r}'_g$ by a left shift of zero more more positions (effectively renumbering the time series indices). From the lagged pair set, we then compute the set of synthetic pairs $(s'_g,s_g)$ where $s'_g$ and $s_g$ are obtained by restricting $\bar{l'}_g$ and $\bar{r}_g$, respectively, to the same maximally overlapping window (that is the same time series indices) of size $w$. Note that in this set $(s'_g,s_g)$ is a possible observation of a causally related pair (modeled after gene $g$), and both $s'_g$ and $s_g$ are times series of size $w$. With our sample hyperparameters this construction will result in time shifts in the index range $\left[0,21\right]$.\footnote{It should be noted that there is an intentional bias in this construction towards synthetic pairs with shorter lag. A more refined model could be developed by using biological assumptions about the lag distribution.} The \textit{full lagged pair set} will be computed correspondingly by random sampling over all genes $g$.

The lagged pairs computed so far account for noise and delay, but do not account for interactions between
genes. To capture interaction, we use a simple equal-weight linear superposition model defined by a single complexity parameter $m$, denoting the number of genes participating in an interaction in addition to the gene of interest $g$. The motivation for using such a simple interaction model, is that typically the Gaussian processes are sufficiently rich (we exploit the high-dimensionality of the time series here) and already exhibit common non-linear behavior, so that additional complexity in the composition would not make a noticeable difference (in addition to the impossibility of estimating detailed interaction parameters from the short time series). Hence, the \textit{synthetic pairs} with interactions are generated by sampling $(s'_0,s_0),\ldots,(s'_m,s_m)$ from the full lagged pair set and computing $(s'_0,s_0 + \cdots + s_m)$. Note how the $s'_1,\ldots,s'_m$ are not used and can also be regarded as additional noise affecting the original causal interaction $s'_0 \rightarrow s_0$. We will see that this additional noisy influence will make causality detection quite challenging depending on the complexity parameter $m$, that we also refer to a \textit{mixin parameter}.

With the method above it is possible to generate positive samples for causal dependency, that are quite realistic and tailored to the specific experimental conditions (e.g. taking into account the variance and noise estimated and represented by the Gaussian processes). Hence, we also refer to the synthetic pair set as the \textit{positive synthetic pair set}. Pure negative samples, on the other hand, are difficult to synthesize from existing expression profiles, because we do not know which pairs of genes are truly independent. Here we make another biologically plausible assumption, namely that the large majority of gene pairs can be considered to be  independent for practical purposes, in other words that the underlying causal network is sparse. With this approximation, we define a \textit{negative synthetic pair set} by sampling $(s'_0,\bullet)$ and independently $(\bullet_0,s_0),\ldots,(\bullet_m,s_m)$ from the synthetic pair set ($\bullet$ stands for a variables that are not used) and computing $(s'_0,s_0 + \cdots + s_m)$. A causal connection between $s'_0$ and $s_0$ can occur by chance, if the underlying genes are dependent, but will be rare under our sparsity assumption, again exploiting the high-dimensionality of our time series.

Finally, positive and negative synthetic pair sets of equal size (e.g. $1000000$ pairs each) are combined into a single balanced \textit{labeled synthetic set} by labeling positive and negative pairs with $1$ and $0$, respectively. Training and validation sets are then generated from this set (e.g. by random split of $90\%$ vs. $10\%$ of the labeled samples).
For comparison purposes, we also generate an \textit{ideal version} of this noisy synthetic data set, by replacing the initial sampling of time series pairs $(\bar{r'}_g,\bar{r}_g)$ from Gaussian processes by sampling just a single time series, in other words, by enforcing $\bar{r'}_g = \bar{r}_g$.

\subsection{Learning Causality using Siamese Networks}

Causality is often associated with a temporal direction, also referred to as the arrow of time, and we would ultimately expect that a causality detector does not only identify the existence of a potential causal connection but also its direction. In our
approach, we decompose the problem by utilizing two models.\footnote{Part of the motivation will become clear later and is related to the practical difficulty of predicting direction even if a causal relationship has already been established.} 
The first model is designed to detect and probabilistically quantify the
existence of causality, while a second  model is used to probabilistically determine the direction under the condition
that a causal dependency is present. One advantage of this two-stage approach is that the two probabilities can be 
clearly separated and the relational properties can be structurally represented in the neural network. 
Our first model will be concerned with a symmetric relation of undirected causality (which is the dual of independence, also a symmetric relation), while the second model will be concerned with an asymmetric relation of directed causality
that can be considered as a temporal refinement.

The \textit{undirected causality detector} is a Siamese neural network, that is a neural network with two identical subnetworks that are each responsible for processing one of the arguments of the binary symmetric causality relation. Each argument is a time series (e.g. of size $w = 80$ consistent with our sample data). The replicated subnetwork is a 1D convolutional network (we use a window size\footnote{The convolutional window size is a hyperparameter that reflects a tradeoff between maximum time lag and maximum size of the patterns that can be taken into account.} $w' = 61$ and stride $1$) with bias and a relu-activation function yielding a tensor  (e.g. $20 \times 50$ dimensional). This is followed by an average pooling layer yielding a vector (e.g. $50$ dimensional), followed by a dense layer with the same output dimension and again with bias and relu-activation. Hence, the output of each subnetwork is a vector in feature space (e.g. $50$ dimensional). The two outputs are combined by a dot-product layer (which captures the symmetry of the problem as part of the architecture). As usual for Siamese networks, the symmetry is also exploited by weight-sharing between the two subnetworks. Finally, to obtain a probabilistic output, a sigmoid function is applied to a linear transformation of the scalar result from the dot-product, which can also be viewed as trivial dense layer with bias and sigmoid-activation.

As a loss function we use binary cross-entropy and as an optimizer we use TensorFlow's implementation of Adam (with default parameters). The key hyperparameters we vary are the type of model (ideal or noisy) and the complexity of the model, using $m = 0, 2, 4, 9$ as settings for the mixin-parameter.  We found that conventional training often gets stuck in low accuracy solutions (around $0.5$ or lower), and while random restarts can help, we opted for a more stable solution based on curriculum learning \cite{curriculum-learning}. To this end, we train the models with increasing complexity of the training data set (we use the mixin-parameter to vary the complexity of the synthetic model). We start with training data generated by $m=0$, which always yields high quality models, and then proceed incrementally to train a sequence of models for $m = 2,4,9$, a process that usually converges to models with similar metrics. Each stage is trained for $1000$ epochs with a batch size of $20000$.

\begin{figure*}[!htb]
	\includegraphics[width=1.0\linewidth]{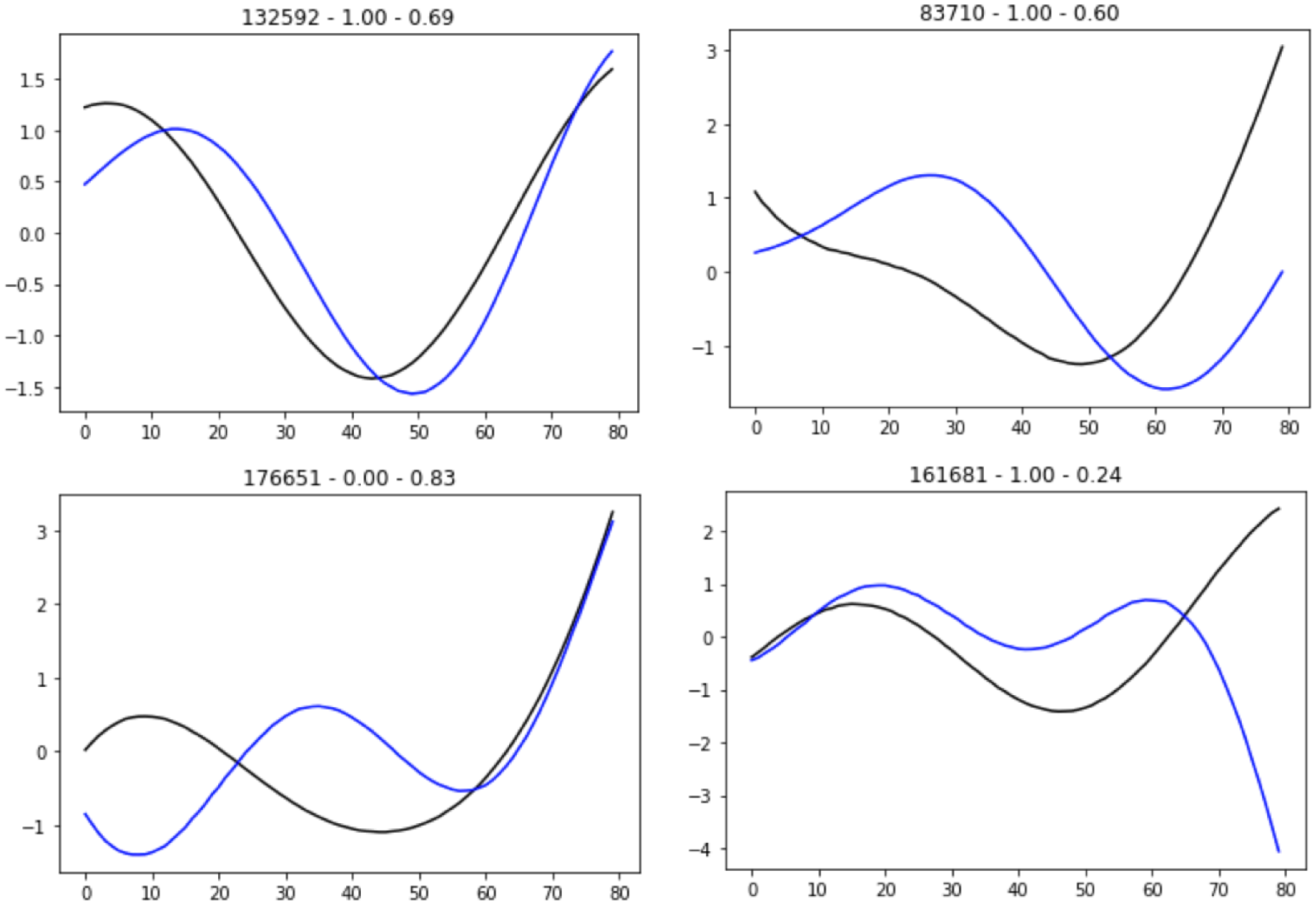}
	\caption{This figure illustrates some of the challenges of causality detection in the presence of noise even with a mixin parameter $m=0$. From left to right the title of each plot contains a synthetic gene pair identifier, the actual probability of causality, and the predicted probability of causality. On the top we see two samples from the test set that are correctly classified as causally dependent (with reasonably high probability). On the bottom we see two incorrectly classified pairs. We have a false positive on the left hand side and a false negative on the right hand side. }
	\vspace{5pt}
	\label{fig-samples}
\end{figure*}

\begin{figure*}[!htb]
	\includegraphics[width=1.0\linewidth]{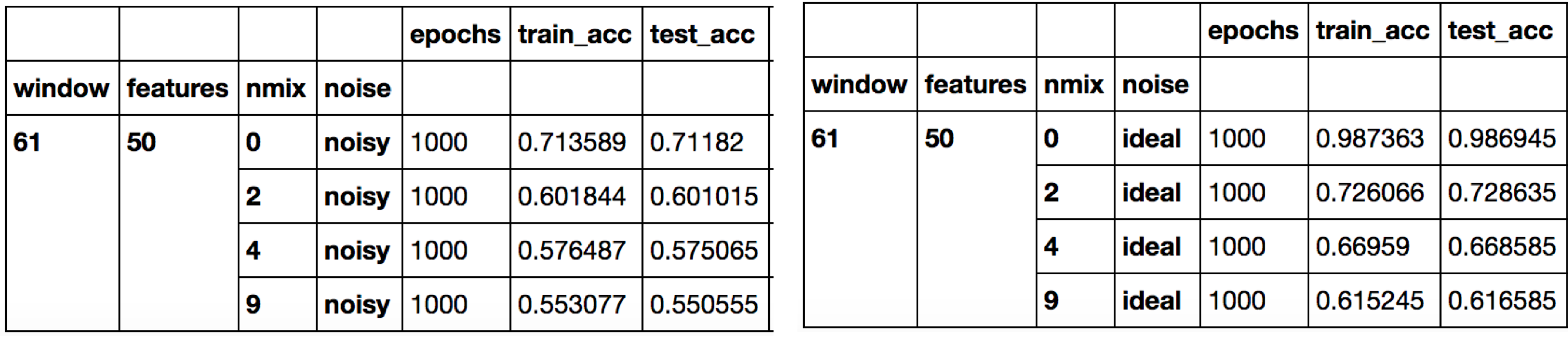}
	\caption{For fixed architectural hyperparameters (convolutional window size and feature dimension), we show
		how the accuracy decreases with increasing mixin parameter that determines the complexity of the synthetic model. For comparison we also show the results for an ideal model without noise.}
	\vspace{5pt}
	\label{fig-results}
\end{figure*}

\begin{figure*}[!htb]
	\begin{centering}
	\includegraphics[width=0.7\linewidth]{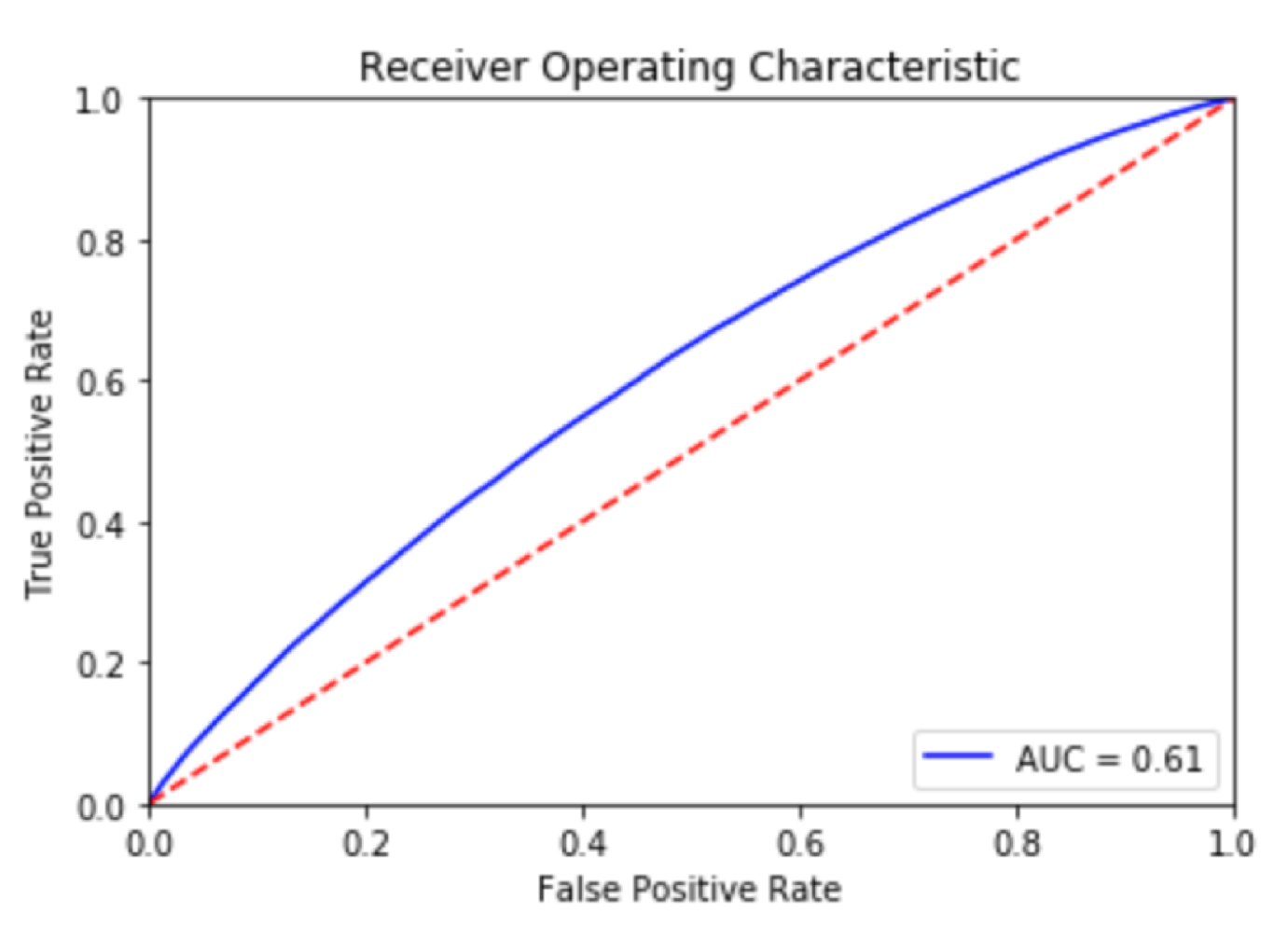}
	\caption{The Receiver Operating Chacteristic (ROC) curve shows the tradeoff between true and false positives (here for the noisy model with $m=4$). For graph synthesis, we restrict our attention to the left part of the curve, keeping the false positive rate below $0.3$. The Area under the Cuve (AUC) is $0.61$, which is another indication of the difficulty of the system identification problem (the quest for completeness) even at the abstract level of causal dependencies.}
	\vspace{5pt}
	\label{fig-roc}
	\end{centering}
\end{figure*}

All models are subsequently validated on the synthetic test sets, which gives results similar to the training set. Some typical
test cases are shown in Fig.~\ref{fig-samples} and clearly illustrate the difficulty of the probabilistic causality detection problem, which cannot be simply reduced to pattern similarity, but needs to be aware of the overall time series distribution and the amount of noise. For one particular experiment ($Unk5$) we show a summary of the test and validation results in terms of accuracy in Fig.~\ref{fig-results}. We can clearly see the increasing difficulty of the causality detection problem with the increasing complexity parameter. While we do not show the results for other RTA experiments, it is worth to point out that the summary results are quite similar.

In Fig.~\ref{fig-roc}, we also show the ROC curve for the model with $m = 4$. In practice, we are interested in higher probability results, so that we use a cutoff of around $0.7$ (corresponding to a false positive rate of $0.3$) at the expense of completeness. Across all RTA experiments, we use probability cutoffs in the range $0.7$ to $0.8$, which still results in detailed causal networks with about a million dependencies. For example, the graphs in Fig.~\ref{fig-biggraph} and Fig.~\ref{fig-graph} that we will discuss subsequently in more detail show only dependencies with a probability of at least $0.7$, in other words, they focus the biologist's attention on the subnetwork for which higher probability relations can be derived, without making any statements about other parts of the network. A more detailed analysis of our validation results shows that higher probabilities are slightly underestimated relative to the synthetic model, which is acceptable for most practical purposes,
but needs to be kept in mind if we approximate an independence relation as the complement of undirected causality.\footnote{This is, for example, relevant in our synthesis of independence graphs, which are interesting in their
own right but not included in this paper.}

\subsection{Extension to Directed Causality}

Determining the direction of causality is the task of another network that is typically applied to
pairs that exhibit a high probability of causal dependency. By estimating the time lag and its direction we embed this decision problem into a continuous problem, for which we can generate training and validation data using a refinement of our synthetic model. As an architecture, we use another  albeit somewhat unconventional type of Siamese network. Its input is again a pair $(g_0,g_1)$ of time series. Its output is not a probability, but simply a real score (trained to be) in the interval $\left[ -1,1 \right]$ and proportional to the temporal lag between $g_0$ and $g_1$. The score $1$ is interpreted as a directed causal dependency $g_0 \rightarrow g_1$ with the highest lag that has been modeled, and $-1$ is interpreted as causality with maximum lag in the opposite direction, that is $g_1 \rightarrow g_0$. A score around $0$ denotes no or a minor lag and hence indecision about the direction (if a causality has already been established). Probabilities will be later associated with the time lag through the validation process.

This \textit{lag detector} is again a Siamese neural network, that is a neural network with two identical subnetworks that are each responsible for processing one of the argument time series. As before, the replicated subnetwork is a 1D convolutional network with bias and a relu-activation function followed by a dense layer again with bias and relu-activation. Unlike the undirected causality detector there is no pooling layer involved. Hence, the output of each subnetwork is a tensor (e.g. $20 \times 50$ dimensional). Weight-sharing is still used between the two subnetworks. After flattening, the two outputs are combined by a subtraction layer (which captures the antisymmetry of the problem) yielding a vector (e.g. of dimension $1000$), The next layer is a dense layer without bias and with tanh-activation (reducing the dimension to $50$), and finally a linear dense layer without bias is used to obtain a scalar output.
As a loss function we use mean square error and as an optimizer we again use TensorFlow's implementation of Adam (with default parameters). For training we use curriculum learning with the same stages and parameters as for the causality detector.

The training and validation data set for the lag detector will be generated in the same way as the (positive) synthetic pair set for the causality detector (e.g. again $1000000$ pairs), but we allow and track positive, negative, and zero time lags in the construction. Furthermore, the labels will be replaced by the $L/M_L$, where $L$ is the lag used in the construction of the pair and $M_L$ is the maximum lag (for our sample hyperparameters we have $L \in [-M_L,M_L]$ with  $M_L = 21$). Training and validation sets are then generated from this set of labeled pairs (e.g. again by random split of $90\%$ vs. $10\%$).

\begin{figure*}[!htb]
	\includegraphics[width=1.0\linewidth]{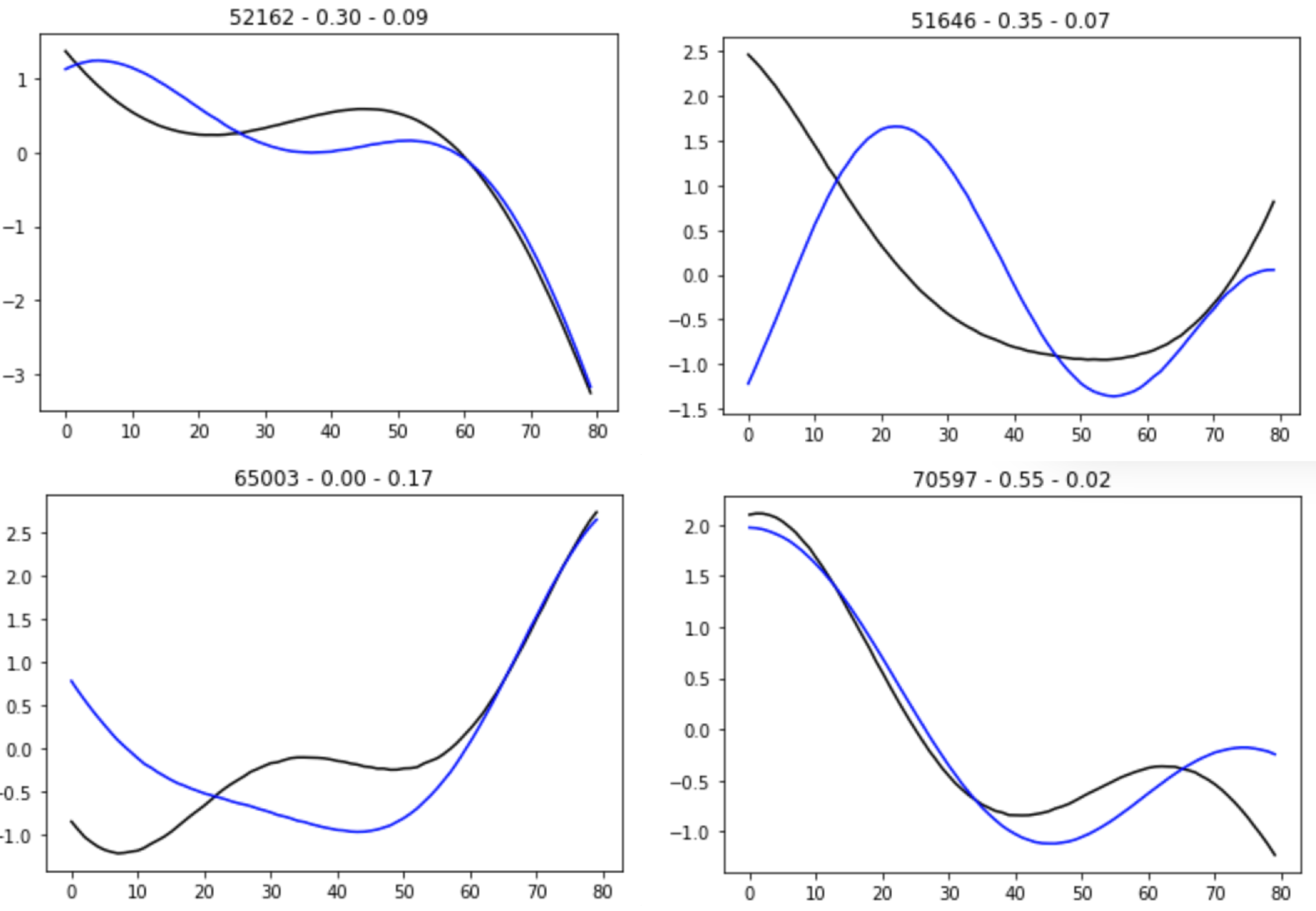}
	\caption{Like causality detection, the quantification of the time lag is a difficult problem due the presence of noise even with a mixin parameter $m=0$. From left to right the title of each plot contains a synthetic gene pair identifier, the actual lag (of the blue relative to the black series), and the predicted lag. On the top we see two samples from the test set where the direction of causality is correctly predicted (but not the magnitude). On the bottom, we see how a delay of zero can lead to a positive predicted lag (left hand side), and a high positive lag leads to a prediction of near zero lag (right hand side).  }
	\vspace{5pt}
	\label{fig-dir-samples}
\end{figure*}

\begin{figure*}[!htb]
	\begin{centering}
	\includegraphics[width=1.0\linewidth]{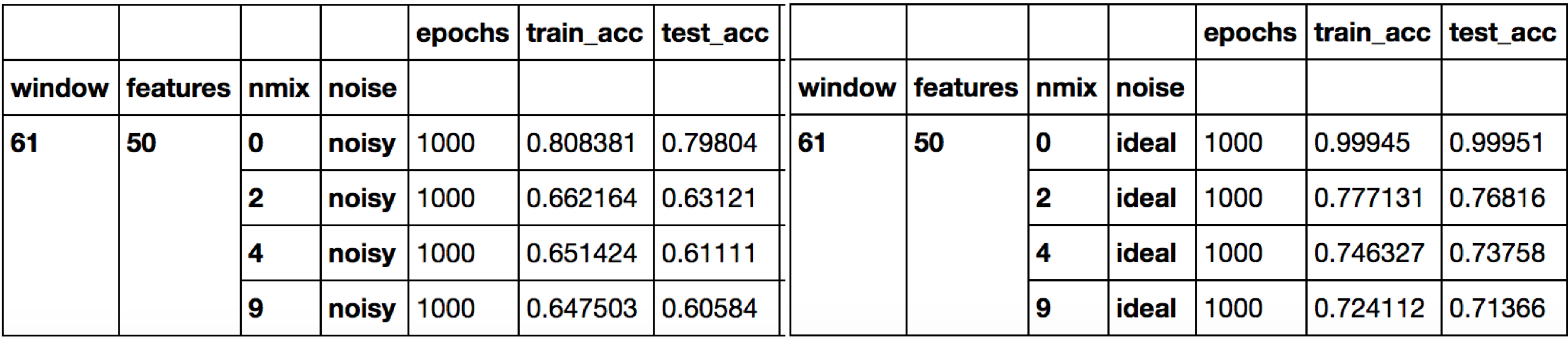}
	\caption{Again we see how the accuracy decreases with increasing mixin parameter that determines the complexity of the synthetic model. The results are based on a fixed threshold $0.025$. The accuracy can be improved by using a higher threshold at the expense of a larger number of pairs for which the direction cannot be determined (precision/recall tradeoff).}
	\vspace{5pt}
	\label{fig-dir-results}
	\end{centering}
\end{figure*}

Sample test cases for the experiment $Unk5$ are again shown in Fig.~\ref{fig-dir-samples} and sample validation results
in terms of accuracy for a fixed lag threshold $0.025$ can be found in Fig.~\ref{fig-dir-results}. For the given threshold,
the test accuracy of $0.61$ (using again the noisy model with $m=4$) is intended for comparison with the other models. 
It may not be sufficient for the biologist, in which case it can be increased by using a larger threshold at the expense of a larger number of pairs for which the direction cannot be decided. Even for a fixed theshold, the test accuracy varies quite a bit across experiments (it ranges between $0.6$ and $0.77$ for the experiments conducted in the RTA project). Another noteworthy observation is the difference between test and train accuracy, which might point to some degree of overfitting in spite of the large amount of training data. However, we found that for other experiments this difference is less pronounced and more likely related to a selected threshold which is too low for the specific experiment. We think that more experimentation is needed to decide if additional forms of regularization would be beneficial for the model.

\subsection{Probabilistic Causal Network  Synthesis}

Once our causality and lag detection models are trained and validated they can be used to synthesize causal networks. In this step, the models are applied directly to the Gaussian processes, more precisely the mean, which is the most likely biological process consistent with the observations, as opposed to samples generated by the synthetic model. Since these graphs can be huge, it is often useful to restrict the genes to a subset, for example using the ranking defined by the 1 or 2 SD filter on Gaussian processes that focuses on genes that are most impacted by the experiment.

\begin{figure*}[!htb]
	\includegraphics[width=1.0\linewidth]{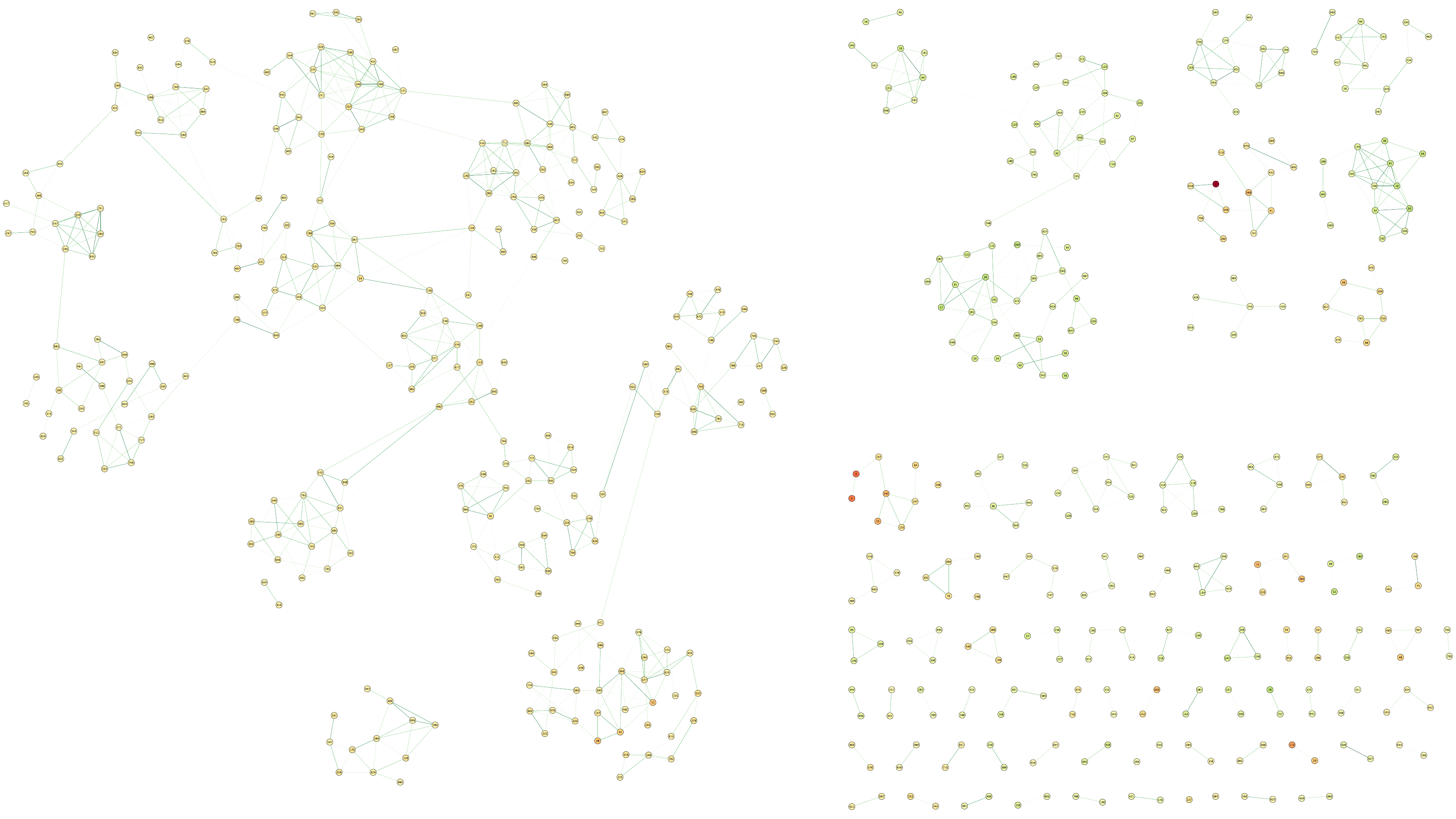}
	\caption{Using a causality detector trained on the synthetic model for experiment $Unk5$ with Gaussian process noise and $m = 4$, we show a graph that depicts for the top 1000 genes passing the 2 SD filter the undirected causality relation  predicted by the causality detector (with a cutoff probability of approx. $0.7$). Genes that are in average up- or down-regulated are colored green and red, respectively. The darkness of the edges indicates the probability of causal relationship. Note that the graph consists of many connected components, including many smaller ones collected in the lower right quadrant. Isolated nodes were removed, because they do not exhibit any high probability causal connections.}
	\vspace{5pt}
	\label{fig-biggraph}
\end{figure*}

\begin{figure*}[!htb]
	\includegraphics[width=1.0\linewidth]{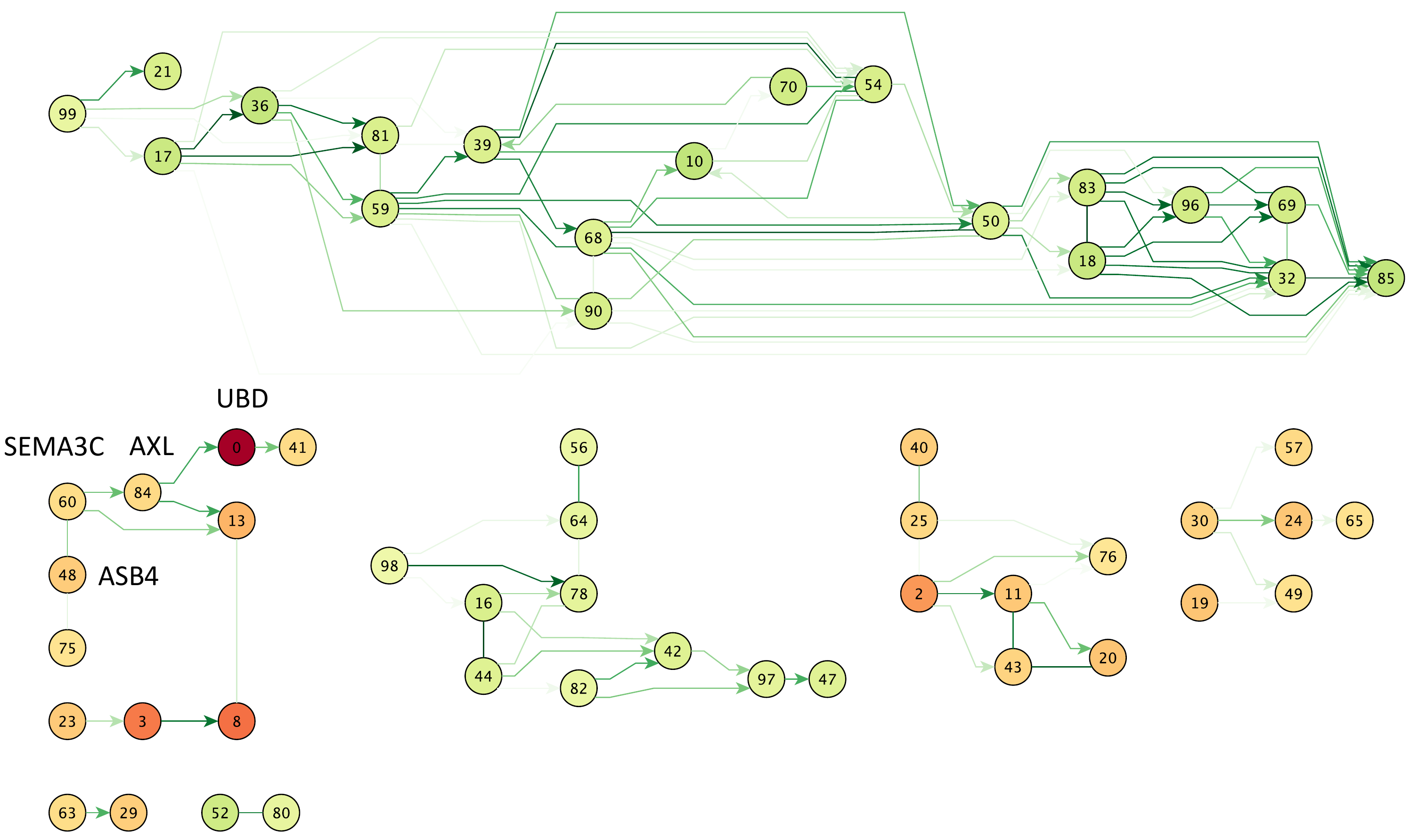}
	\caption{Based on a synthetic model for $Unk5$ with Gaussian process noise and $m = 4$, we show a graph that combines the undirected causality relation predicted by the causality detector (with a cutoff probability of approx. $0.7$) and the direction predicted by the lag detector (with a threshold $0.025$ corresponding to a probability of approx. $0.6$). Undirected edges show that the direction cannot be decided with sufficient probability. For clarity, the graph is restricted to the top 100 genes passing the 2 SD filter. The large upregulated component on the top contains many genes related to cholesterol regulation. Four down-regulated genes corresponding to our sample Gaussian processes from Fig.~\ref{fig-gp} are labeled and involve the strongly down-regulated UBD. It is interesting to note that no sufficiently high probability direction can be established between ASB4 and SEM3C, even if the direction might seem obvious
		from visual inspection.}
	\vspace{5pt}
	\label{fig-graph}
\end{figure*}

An \textit{undirected causal network} is defined by nodes corresponding to all genes in given subset and edges between pairs of genes for which the causality detector detects a dependency with at least the cutoff probability, which is another parameter in the graph synthesis process. In our automated workflow, we vary it between 100 and 1000 to generate graphs that are interpretable by biologists. An example of such a graph is shown in Fig.~\ref{fig-biggraph}.

A \textit{directed causal network} is a refinement of an undirected causal network. Each edge is directed according to the
prediction of the lag detector based on a positive threshold in the interval $(0,1]$ that was associated with a probability during validation. Each undirected edge becomes a directed edge if the lag prediction reaches at least the positive threshold and remains undirected otherwise. An example of such a graph in shown in Fig.~\ref{fig-graph}.

\subsection{Biological Plausibility Check}

In this section, we argue why our simple equal-weight Gaussian process mixture model and especially the use of a mixin complexity parameter between  $2$ and $9$ is biologically plausible and a reasonable tradeoff given the limited amount of data available. The limitation comes mainly from the short duration of a biological experiment, which is unavoidable, because cells have a very limited lifetime outside of the human body leading to an overlay with a multitude of other processes not related to the experiment and furthermore the impact of a system perturbation naturally exhausts itself as time progresses. 

\begin{figure*}[!htb]
	\begin{centering}
	\includegraphics[width=0.7\linewidth]{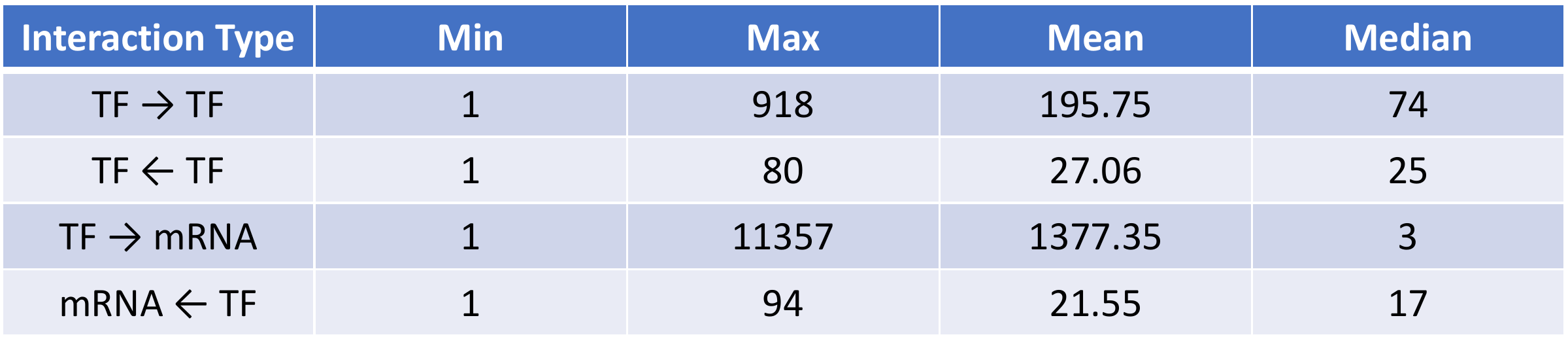}
	\caption{Statistics of degree distribution for various interactions in the integrated human gene regulatory network of  the ENCODE project \cite{Gerstein12}.}
	\vspace{5pt}
	\label{fig-encode}
	\end{centering}
\end{figure*}

To understand the complexity of a typical individual MoA in the context of transcriptomics, we recall that we view the 
MoA as a process and hence a subnetwork of the gene regulatory network, which should give us an estimate for
the upper bound complexity. The richest source currently available for the human gene regulatory network is the ENCODE project \cite{Gerstein12}, which identified interactions between transcription factors (TF), micro-RNA (miRNA), and
all remaining messenger RNA (mRNA). In the RTA project, we are not concerned with miRNA, although it is a factor contributing to the biological variation. The degree distributions of the proposed integrated network are shown in Fig.~\ref{fig-encode}. To use a robust statistics, we are mostly interested in the median in-degree of genes (mRNA and TF, which we do not distinguish a priori) which is between $17$ (for the TF/mRNA interactions) and $25$ (for TF/TF interactions which are much more rare with less than $10\%$). Knowing that, first not all transcription factors are equally weighted, and second that it is a complex combination of transcription factors that typically controls transcription, the impact of each single transcription factor is greatly diminished especially in the context of a very specific experiment. We conservatively estimate that this should lead to reduction of at least a factor of $2-7$ that should bring us to an in-degree in the range of $3-10$ and hence a complexity parameter in the range of $2-9$.

Previously we have used $m = 4$ for causal graph synthesis. To understand if this choice is at least biologically plausible, we use the Connectivity Map \cite{Lamb06} data to establish an approximate ground truth for undirected causality based on correlation (with all its known limitations). To this end, we use the early version of the raw Connectivity Map data based on complete transcriptomics via microarrays (albeit an older chip generatation compared to the one we use in the RTA project). The data contains signatures arising from the perturbations of 5 different cell lines with more than $1000$ small molecules in 6100 experiments. Using a conservative assumption that this set is diverse enough to contain at least $500$ independent experiments, we can compute the Pearson correlation between each pair of genes across all 6100 perturbations, and use a threshold of $0.075$ to establish a set of known (positively) dependent gene pairs as a ground truth with a p-value below $0.1$.  We also establish a set of $11261$ common (protein-coding) genes that were observed in both projects.

\begin{figure*}[!htb]
	\begin{centering}
		\includegraphics[width=0.7\linewidth]{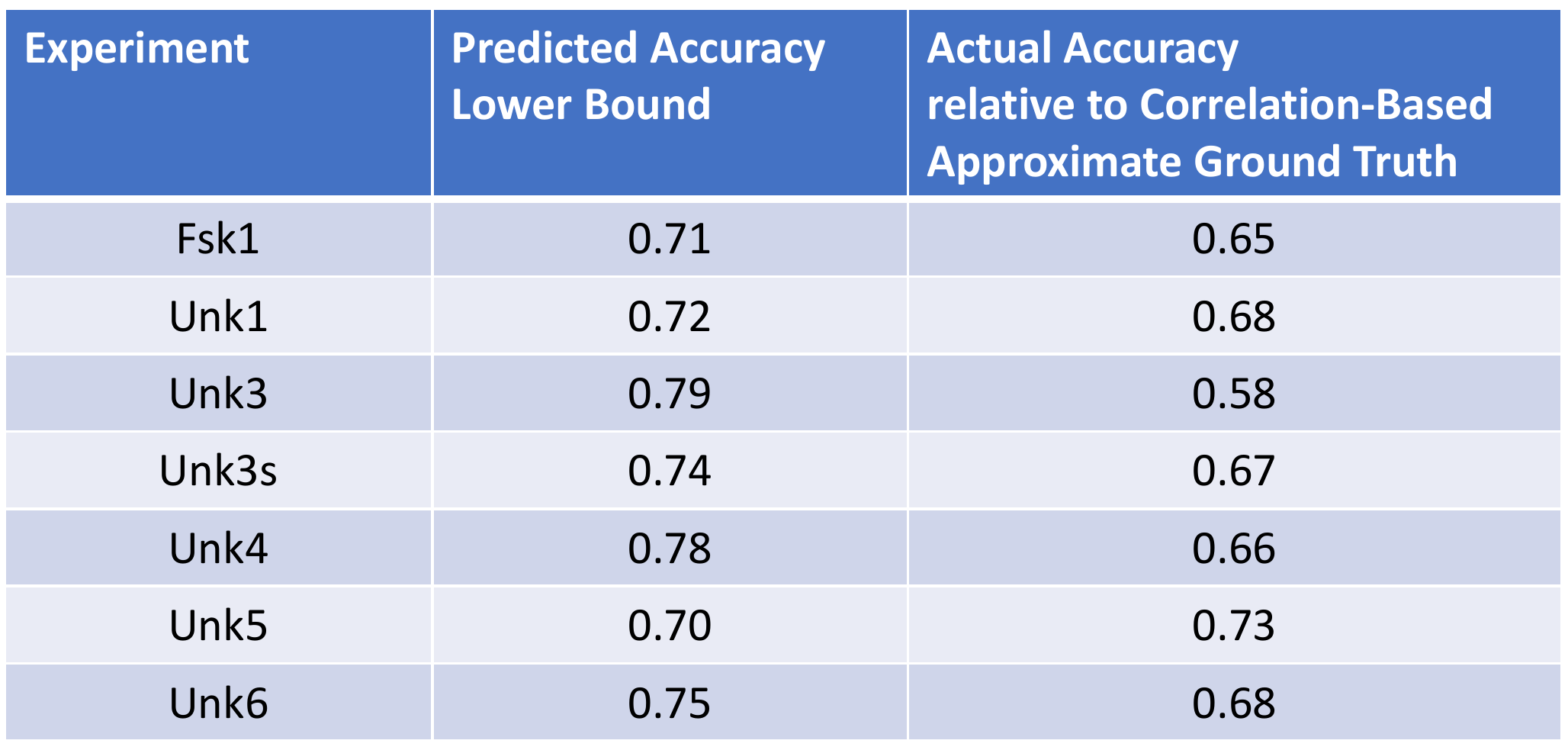}
		\caption{Predicted accuracy of large undirected causal networks for different RTA experiments is compared with actual accuracy relative to approximate ground truth based on Connectivity Map correlations, which is naturally incomplete. All results are for a fixed mixin parameter $m = 4$ and synthesized undirected causal network graphs with close to $1M$ edges.}
		\vspace{5pt}
		\label{fig-plaus}
	\end{centering}
\end{figure*}

The number of undirected pairs in the correlation-based dependency set is $35500931$ which reduces to $32741966$ pairs of common genes (about $51\%$ of all possible pairs) that we also observe. The corresponding graph will serve as approximate ground truth. We then use our undirected causal graph synthesis to generate a large graph with $950078$ edges (no filtering applied), which is still possible with a probabilistic lower bound of $0.7$ for causal dependency in case of $Unk5$. The graph is then also reduced by restricting the nodes to the set of common genes yielding $487235$ edges. Both of these graphs are clearly not for human consumption, but enable us to do some meaningful statistical analysis of the accuracy of the later relative to the former: $354765$ of the $487235$ edges, that is about $73\%$ are also in the approximate ground truth dependency graph. While these results for experiment $Unk5$ are approximately consistent, it turns out that other RTA experiments show larger differences as summarized in Fig.~\ref{fig-plaus}. In some cases, outliers might indicate experimental problems (e.g. in case of $Unk3$). The variation across experiments, however, might also point to the limitations in the ground truth (e.g. the bias towards certain types of perturbations) and in addition to the natural incompleteness of correlation-based causality. Another possible but in our view less likely explanation is that the typical behavior is more complex than captured by our mixin parameter $m = 4$, potentially suggesting the use of larger complexity parameters, which also have been explored in the RTA project.

\section{Heuristic Causal Network Synthesis}

In this section, we briefly present two other causal network synthesis approaches that we developed in the RTA program. The first one is based on convolutional autoencoders and implements a novel evidence-based approach to detect and quantify inconsistencies in the model using a quite general form of domain knowledge. It is an instance of a more general approach that uses domain knowledge for validation and model selection, which should be contrasted with our more recent work \cite{palo}, where domain knowledge is directly incorporated into the neural network model. With the second network
synthesis approach presented below we explore the feasibility of applying deep neural networks in this area, which is
not obvious given the short duration of observation that limits the amount of data that can be collected. We therefore
explore an adaptive approach parameterized by complexity that includes a range of deep and wide neural networks as part of a broader model family. Both of these approaches to network synthesis are very different from our Siamese neural network approach and hence can serve as complementary source for biological hypothesis generation. The price for their higher complexity is that these methods do not provide a probabilistic interpretation of the resulting networks, but on the
other hand the results do not depend on a specific synthetic model and hence are equally valid for all types of data
collected in the RTA project.

\subsection{Synthesis based on  Convolutional Autoencoders}

Different types of autoencoders \cite{autoencoders} are already used in the RTA workflow for the detection of anomalies \cite{Vertes18}. Here we apply them to detect potential causality. Like our Siamese networks, one motivation for this approach is the need to detect more complex causal relationships that go beyond the type that can be detected by cross-correlation, which is at the core of the baseline algorithms in our RTA workflow \cite{Vertes18}. Convolutional autoencoders \cite{conv-autoencoders} have been successfully used in image processing due to their capability to learn reoccurring patterns and perform a dimensionality reduction in an unsupervised setting. In logical terms, they structurally encode the homogeneity of the input space, a type of symmetry that can be equationally defined by translation invariance.

In a nutshell, we train a family of 1D convolutional autoencoders (with average pooling and elu-activation) to learn common patterns that are distributed over the time series (not occurring only at a particular time). Hyperparameters are used for feature dimension (e.g. various settings between $3$ and  $100$) and window size (e.g. $31$, $41$, $51$, $61$ out of $101$ time points). The training set consists of $90\%$ of the log-ratio time series set estimated using our Gaussian process model. The remaining $10\%$ are used for validation for autoencoder performance. The autoencoders
are trained with a mean square error loss using TensorFlow's Adam optimizer with default parameters and an early stopping rule to reduce overfitting. The resulting encoders transform each window into a lower dimensional feature space (between $3$ and $100$ dimensions). In the next phase, the convolutional subnetwork of the encoder will be used as a standalone network to quantify the expression of the features over time (e.g. for $60$ different windows of the full time series if the window size is $41$) for all estimated Gaussian processes (the mean time series) of a given experiment.

We now define a \textit{feature occurrence} as a triple consisting of the feature vector and the gene and time point (the center of the feature/pattern) where it was detected. To identify similar features, we compute the pairwise Euclidean distances between the features of all such occurrences. Using a variable distance threshold hyperparameter, we obtain a reduced set of pairs of such occurrences representing \textit{approximate matches} (across genes and time points). We also say that each such match is a \textit{witness} for potential causality. A pair $(o, o')$ of two occurrences is \textit{directed}, written $o \rightarrow o'$, if the time associated with $o$ is strictly lower than the time of $o'$. It is \textit{undirected} and written as $o \leftrightarrow o'$ if the associated times are equal (that is not distinguishable within our finite time resolution).

 In this evidence-based (rather than probabilistic) approach, we introduce another hyperparameter $n$, the number of witnesses (we use settings $1$, $2$, $3$, $5$, and $10$), to lift the causal relation from feature occurrences to genes: A pair of genes $(g,g')$ is in the (symmetric) \textit{undirected causality} relation, written $g \leftrightarrow g'$, iff there are at least $n$ witnesses of undirected causality involving $g$ and $g'$. A pair of genes $(g,g')$ is in the \textit{directed causality} relation, written $g \rightarrow g'$, iff  there are at least $n$ witnesses of directed causality involving $g$ and $g'$ in this order. Immediate inconsistencies are removed from this directed relation by requiring that it does not contain pairs of undirected causality or pairs of directed causality in the opposite direction.

\begin{figure*}[!htb]
	\begin{centering}
		\includegraphics[width=\linewidth]{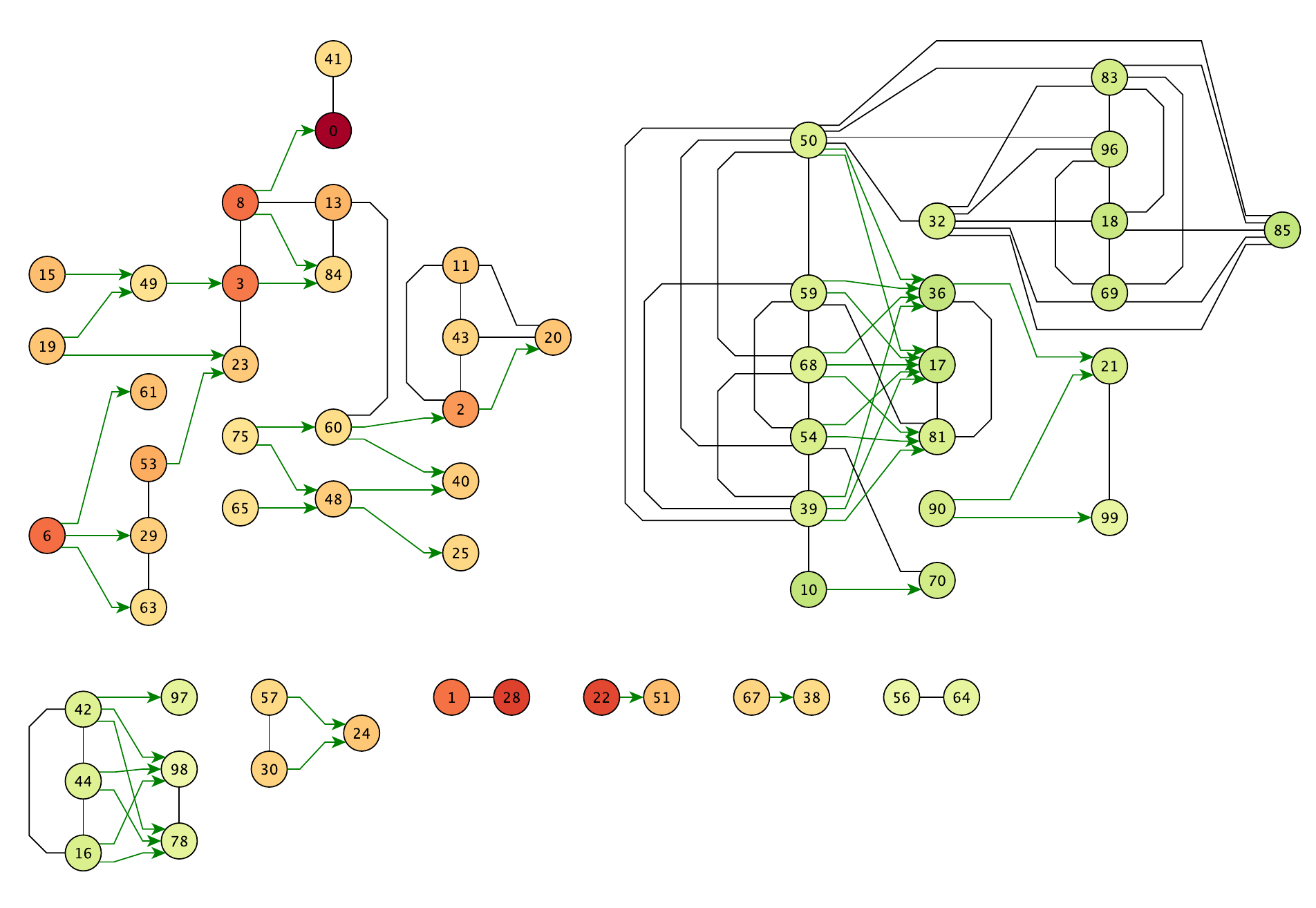}
		\caption{Autoencoder-Based Network Synthesis. We show the synthesized causal network corresponding to the top model for $Unk5$ for window size $61$ (which uses a modest feature dimension $5$ and $5$ witnesses) limited to the top $100$ genes passing the 2 SD filter. Directed edges denote potential causality. Undirected edges denote a potential causal connection for which the direction cannot be determined. As usual, isolated nodes are eliminated from the graph. Note the large fraction of undirected edges in the large green component, again related to the Cholesterol pathway. Often switching to a model with smaller window size can resolve the direction but at the cost of higher uncertainty.}
		\vspace{5pt}
		\label{fig-autoenc-graph}
	\end{centering}
\end{figure*}

\begin{figure*}[!htb]
	\begin{centering}
		\includegraphics[width=\linewidth]{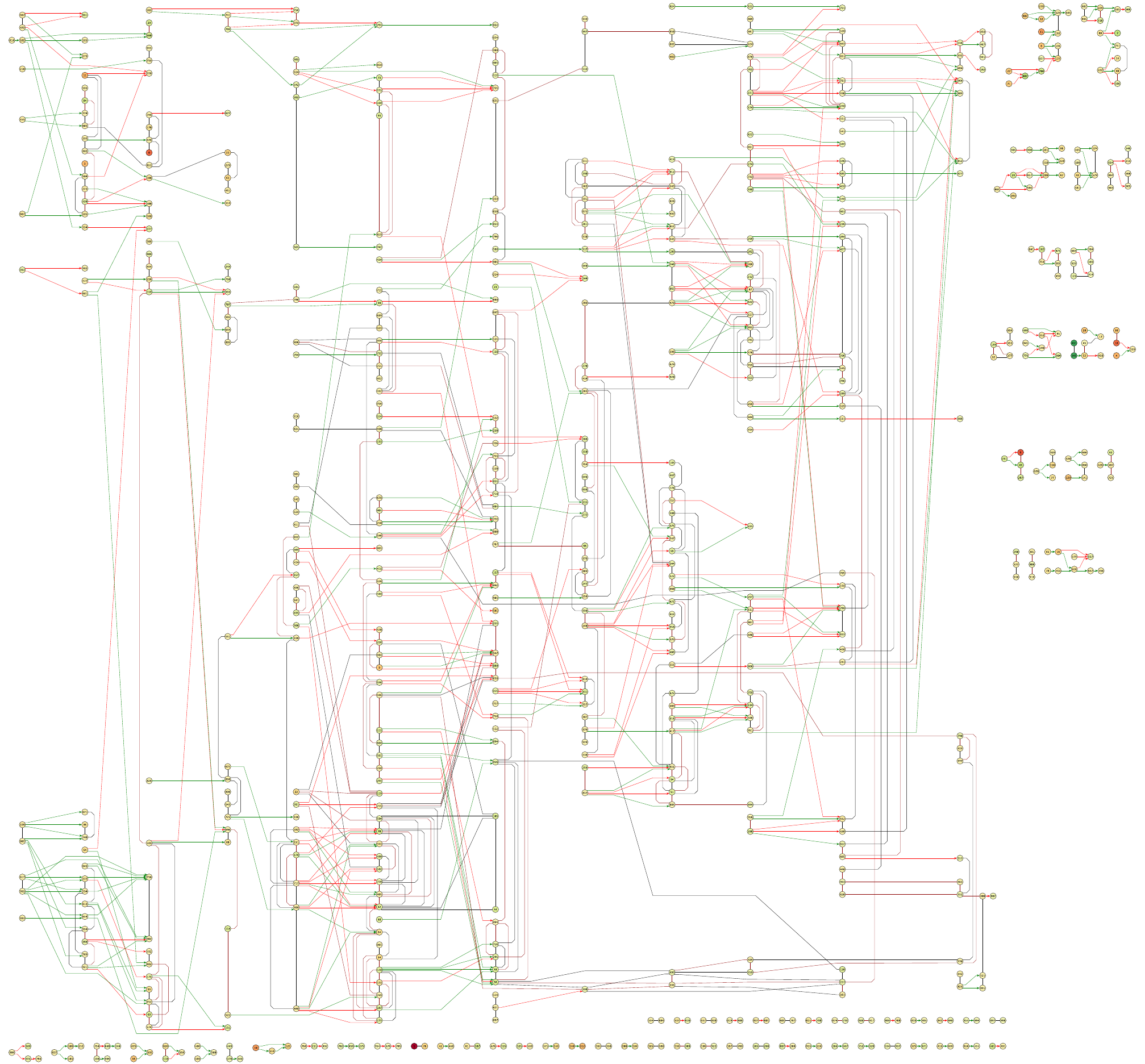}
		\caption{Integrated Autoencoder-Based Network Synthesis.  An integrated causal network corresponding to the top model for $Unk5$ for window size $61$ (which uses a modest feature dimension $15$ and only $2$ witnesses) limited to the top $1000$ compounds passing the 1 SD filter. Based on a generalized algorithm, it includes positive (green) and negative (red) causality. Most edges are directed from left to right. Dark edges are undirected. The graph contains genes and proteins, and location information for some of the proteins that is not visible here (nucleus vs.\ cytoplasma). Many smaller disconnected components are shown on the side.}
		\vspace{5pt}
		\label{fig-autoenc-graph-int}
	\end{centering}
\end{figure*}

The distance threshold hyperparameter is computed to target a specific causal network size that is suitable for
exploration and interpretation by the biologist. We usually generate multiple graphs with different sizes in the 
range of $100$ to $1000$ edges and let the user select the appropriate level of detail and switch between different views. Overall we obtain a family of causal networks (representing both directed and undirected causality between genes) parameterized by convolutional window size, feature dimension, number of witnesses, and network size. Finally,  generic domain knowledge about the nature of causality (not specific to the biological domain) is used to heuristically quantify the consistency of the witnesses (utilizing probabilistic lower bounds for inconsistencies) and hence the plausibility of these models. The results are then combined with validated model performance to establish a ranking of all models. For each window size, the three top models in this ranking are actually generated in the form of graphs and can be further explored by the biologist. A sample model is shown in Fig.~\ref{fig-autoenc-graph}.

The advantage of our autoencoder-based graph synthesis algorithm is that very few assumptions are needed for this evidence-based approach. In the context of the RTA workflow, they are not only applied to transcriptomics data sets across all experiments but also to proteomics and metabolomics data, resulting in integrated causal networks that can include nodes and potential causal relations for all three types of biological compounds. The RTA workflow  also includes a generalized algorithm that captures positive and negative (that is inhibitory) causal relationships. A sample result is shown in Fig.~\ref{fig-autoenc-graph-int}.

\clearpage

\subsection{Synthesis with Deep and Wide Neural Networks}

With the rapidly increasing computational capabilities and the related progress in deep learning applications it is natural to explore if more complex models that might reveal additional information about the nature of causal interactions can be
synthesized and meaningfully validated in spite of the relative short duration of our time series. Since it is not
a priori clear what the appropriate model complexity is given the amount and nature of the data, we use a parameterized family including a range of deep and wide neural networks and perform a comparative model validation. 

To enable insight into the nature of causal interactions, we consider a high-dimensional predictive setting, where based on the changes exhibited by the log-ratio time series between two time points $t_{i-1}$ and $t_{i}$ of our Gaussian processes (using again $100$ time points for the $48h$ period) the neural networks will be trained to predict the changes for all genes from $t_{i}$ to $t_{i+1}$.  Each change is represented as an approx. $17K$-dimensional vector with a component for each protein-coding gene. For a fixed experiment, the sample set is the data set of all pairs of a change vector and the label, which is the change vector for the next time point. We use the first $90\%$ of the $48h$ time scale for training
and the remaining $10\%$ for validation, with the idea that predicting the long-term effect will be most challenging
and biologically relevant (in spite of some bias introduced by this choice).

We use depth $d$ and width $w$ as architectural hyperparameters ranging over natural numbers in  $[2,10]$ and $\set{2^i \textrm{ for } i \in [2,12] }$, respectively. A \textit{deep and wide neural network} is then defined by stacking $d$ composite layers each consisting of a dense layer of width $w$ without bias but with $L_1$-regularization (using a coefficient $0.00000001$), followed by a drop out layer (using a low dropout rate of $0.05$), and followed by an elu-activation layer. Input and output dimension is $n$, the number of protein-coding genes (approx. $17K$). Dropout and regularization
parameters were experimentally determined to reduce overfitting, which is easily possible given the huge
number of network parameters, especially for networks with depth $10$ and width $4096$. As a loss function we use mean square error (MSE) and we again use TensorFlow's implementation of Adam (with default parameters). We train each network for $1000$ epochs with a batch size of $10$.

\begin{figure*}[!htb]
	\begin{centering}
		\includegraphics[width=0.7\linewidth]{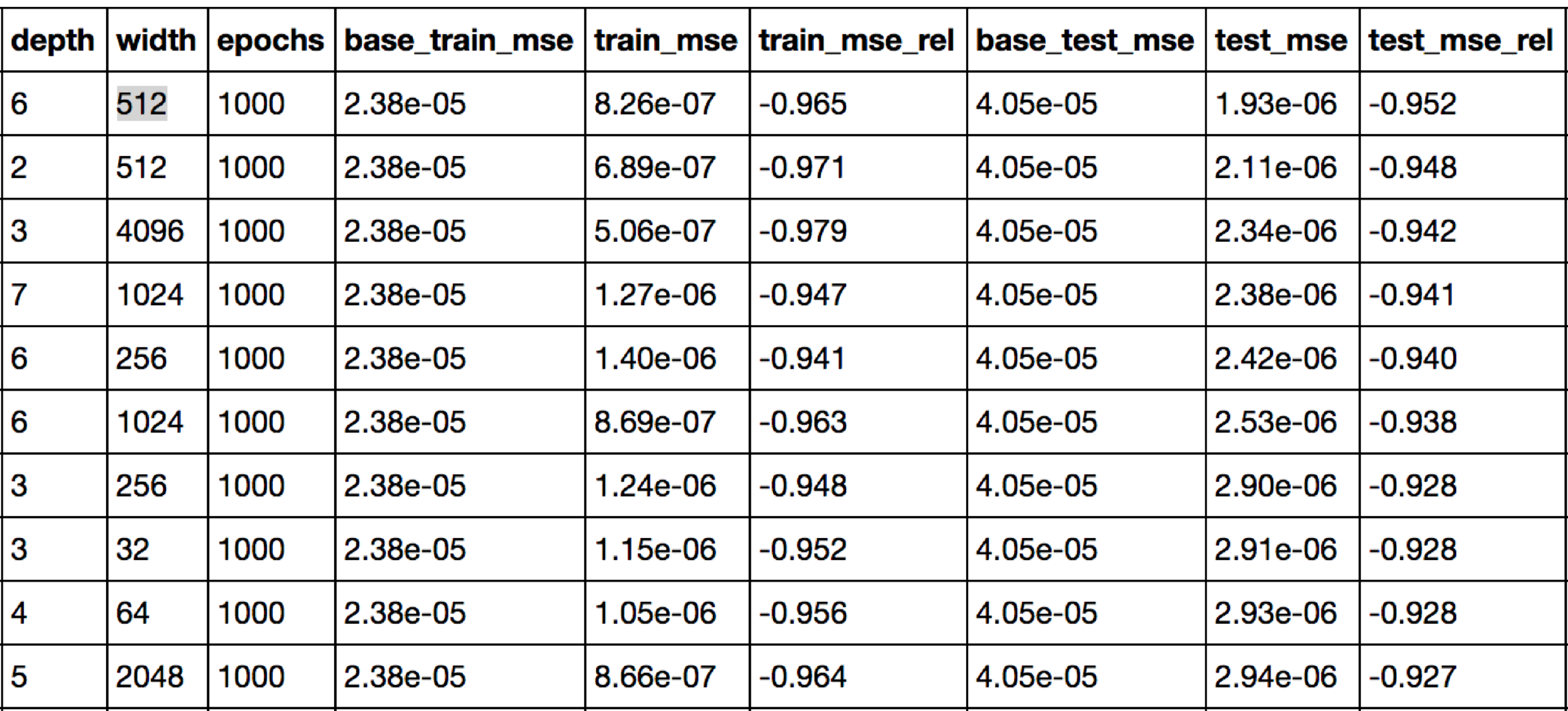}
		\caption{Comparative model validation of deep and wide neural networks for the RTA experiment $Unk5$. We only show the top 10 models ranked by relative test MSE (last column). It is noteworthy that the top model is a deep and wide network with depth $6$. The next two models have much smaller depth $2$ and $3$, but their predictive performance is worse, which at first seems to indicate that the dynamics of $Unk5$ requires a fairly complex explanation. }
		\vspace{5pt}
		\label{fig-deepwide}
	\end{centering}
\end{figure*}

\begin{figure*}[!htb]
	\begin{centering}
		\includegraphics[width=0.7\linewidth]{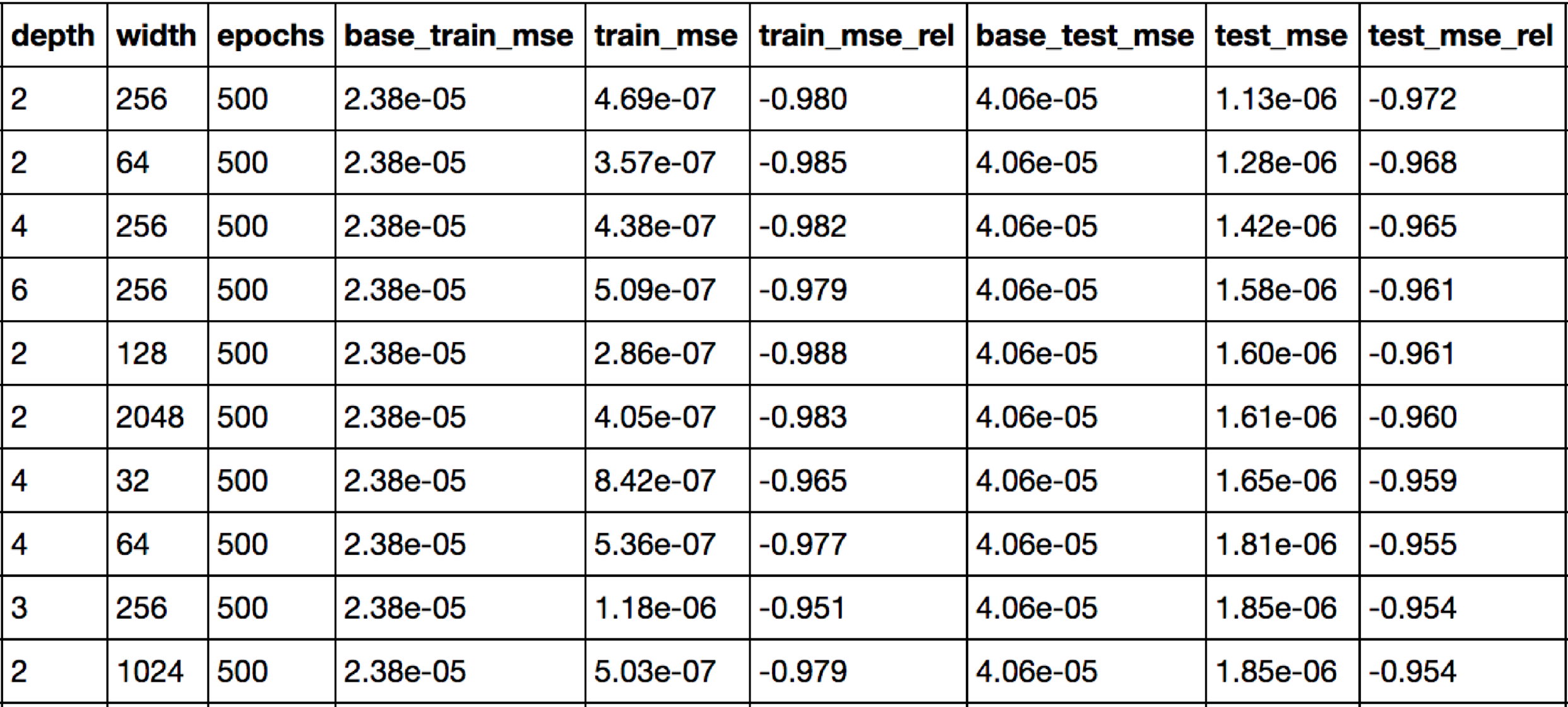}
		\caption{Comparative model validation of deep and wide neural networks for the RTA experiment $Unk5$ with pretraining on all transciptomics data sets of the RTA project. Again we only show the top 10 models ranked by relative test MSE. It is noteworthy that the top models are networks of depth $2$, indicating that in the context of
			all data relatively low complexity models provide the best explanation for $Unk5$, but a higher complexity model
			with depth $6$ is still among the top $10$ and can be used if more detail is needed. It also has better relative MSE
			and lower complexity than the top model without pretraining.}
		\vspace{5pt}
		\label{fig-deepwide-pre}
	\end{centering}
\end{figure*}

Since we are concerned with populations of biological cells, changes in the obsvervables do not occur abruptly (a property that is already reflected in our Gaussian process model). Hence we measure the improvement in MSE relative to the default prediction that the change is the same at the next time point (persistence property). A summary of validation results for experiment $Unk5$ is shown in Fig.~\ref{fig-deepwide}. While the difference between test and training MSE points to some degree of overfitting, it is not severe in terms of relative MSE. Generally, we observe a high degree of predictability in the data set, which is consistent across all experiments. For example, the top model in Fig.~\ref{fig-deepwide} has a $95.2\%$ lower MSE than the default prediction.

A natural question is if the models can be further improved by using more data even if it is not directly related
to the experiment in question. To test this idea, we added a pretraining stage to our workflow, where all the neural networks
are trained on all transcriptomics data sets generated in the RTA project (but only for $500$ epochs). In the second
stage, we then continue the training using only data of the experiment for which the causal network needs to be constructed (again only for $500$ instead of $1000$ epochs). Sample results for $Unk5$ can be found in Fig.~\ref{fig-deepwide-pre} and indicate further improvements. In fact, improvements in relative test MSE 
were obtained for most RTA experiments, except for two cases where the numbers remained approximately the same.

\begin{figure*}[!htb]
	\begin{centering}
		\includegraphics[width=\linewidth]{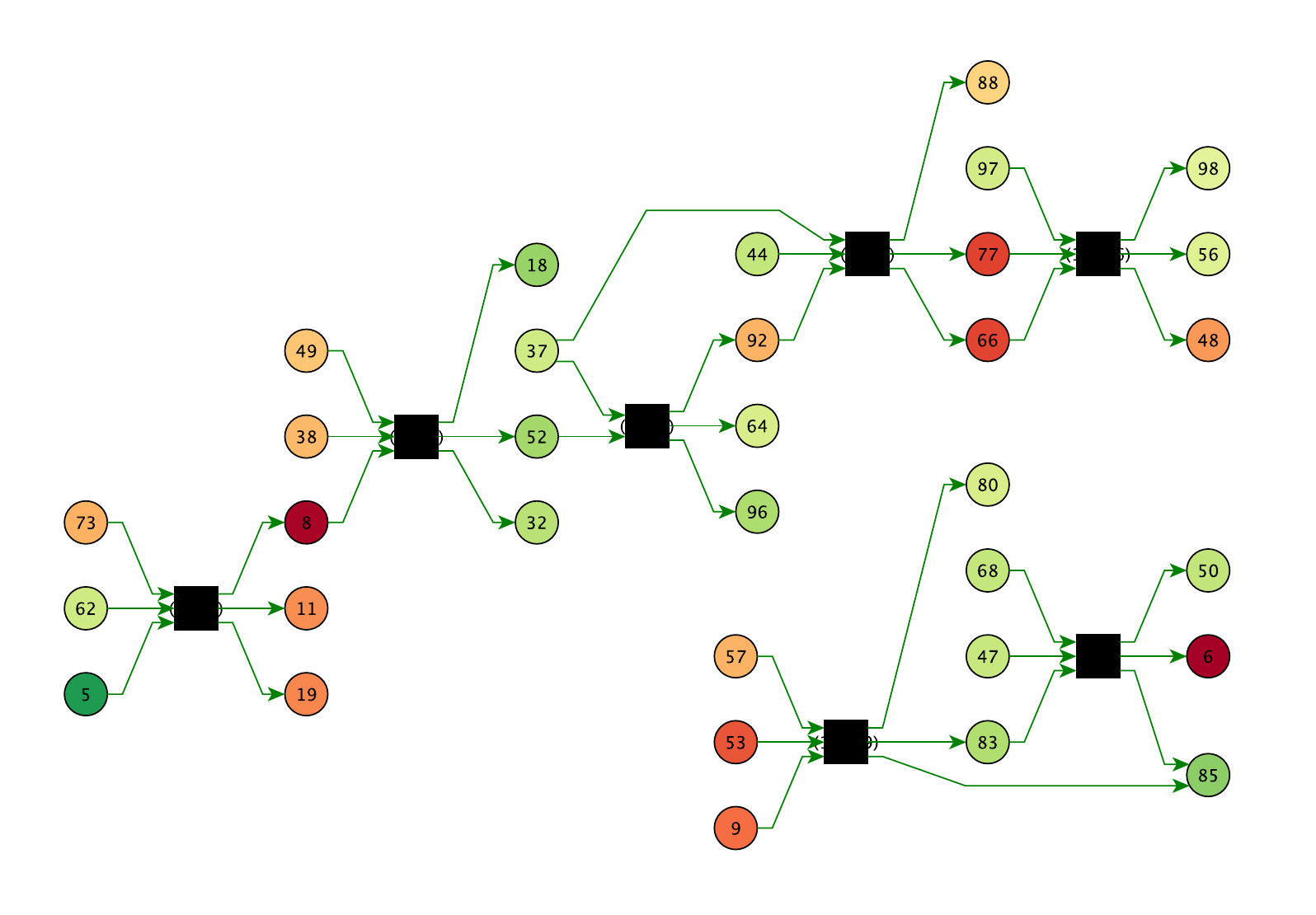}
		\caption{The network corresponding the the top model for $Unk5$ with pretraining but limited to the top $100$
		genes passing the 2 SD filter and limited to a maximum degree of $3$ for a concise presentation.}
		\vspace{5pt}
		\label{fig-deepwide-pre-net}
	\end{centering}
\end{figure*}

\begin{figure*}[!htb]
	\begin{centering}
		\includegraphics[width=\linewidth]{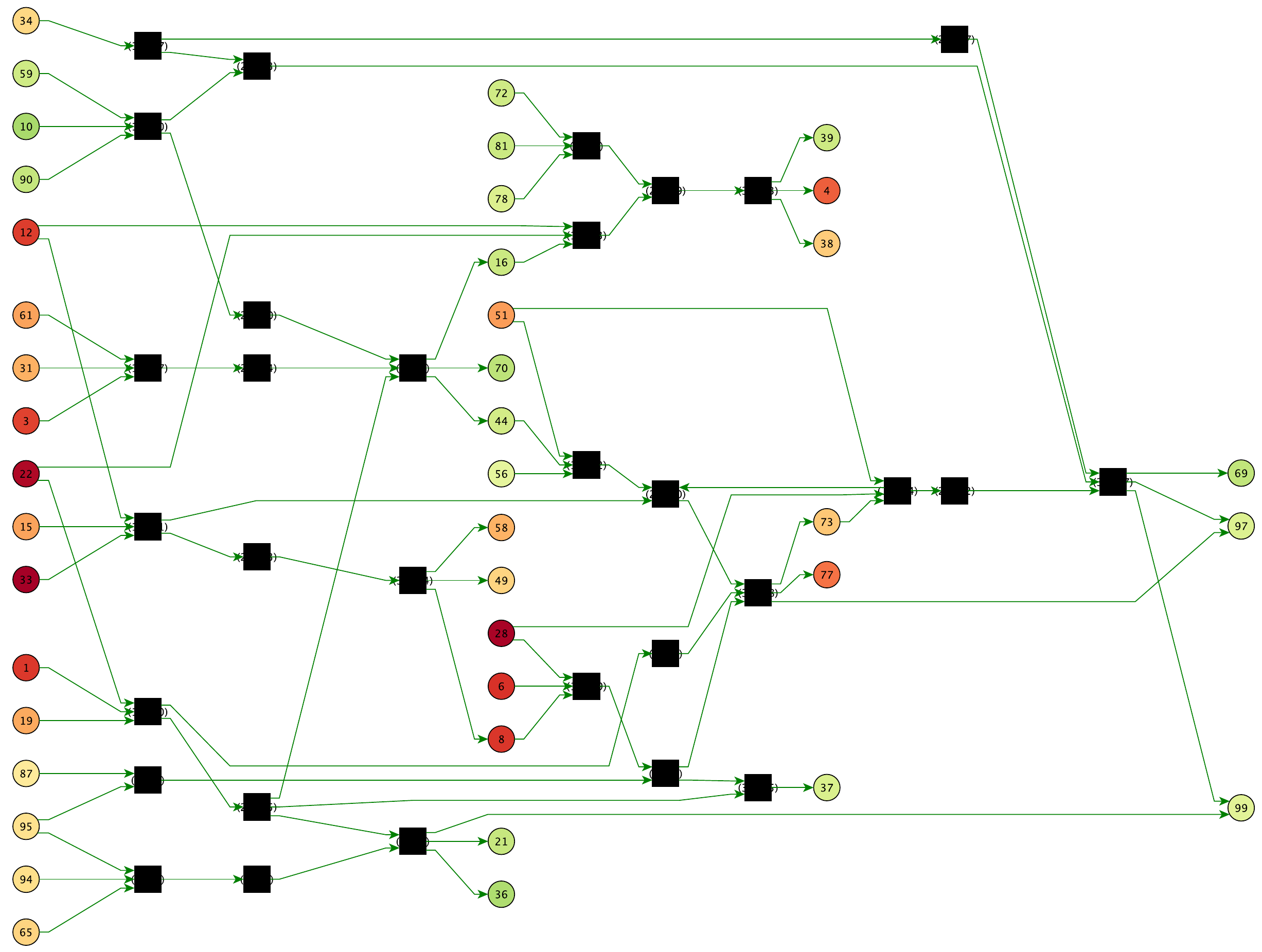}
		\caption{The network corresponding the the best depth 4 model for $Unk5$ with pretraining but limited to the top $100$ genes passing the 2 SD filter and limited to a maximum degree of $3$ for a concise presentation.}
		\vspace{5pt}
		\label{fig-deepwide-pre-net2}
	\end{centering}
\end{figure*}

When evaluating the validation results across all experiments we can observe that with pretraining the top models
are always of depth $2$ (or $3$ in one case), which seems to confirm that pretraining has an effect similar to regularization or incorporation of background knowledge in reducing model complexity. In most cases, however, deep networks
are still included among the top 10 ranked models (for which we generate graphs), and a biologist might prefer them because they provide an alternative or more detailed view, albeit at the cost of slightly lower predictive accuracy.

Compared with the earlier approaches, our causal network synthesis is more heuristic for deep and wide neural networks
that have been trained and validated as explained above, and primarily driven by the need to obtain graphs of manageable complexity. In essence, we use the neural network structure as the template for our causal network, or in other words the causal network is simply a (very) abstract visualization of the structure of a deep and wide neural network. The key here is to focus on the most important subnetwork, which may depend on the level of detail the user is interested in. Hence we generate a \textit{parameterized family of directed causality graphs} with two types of nodes. The parameters are the common maximum in- and out-degree (we use settings $3$, $5$ and $10$) and the maximum number of genes to consider (for which we use $100$, $200$, and $500$). For a given subset of genes (which is typically a subset defined by our SD ranking satisfying the maximum bound), we then construct the graph by the following multistep process. Initially, the nodes of the graph are simply all units of the neural network. Inputs and output units are identified with their corresponding genes. Each positive-weight connection in the neural network can potentially give rise to an edge, but we determine a suitable weight threshold for each layer so that edges can be defined as the subset of neural connections with a weight above the threshold. We choose the lowest threshold that still leads to a number of edges bounded by the square root twice the number of possible connections (a sparsity constraint). In the next step we perform layer-wise edge pruning (starting with edges derived from lowest weight connections within each layer) till the in- and out-degree for each node satisfies the required bound. Examples of synthesized networks can be found in Figs.~\ref{fig-deepwide-pre-net} and \ref{fig-deepwide-pre-net2}.

Variations of this algorithms are clearly possible, e.g. by varying the sparsity constraint. Also if we take into account positive and negative weights, we obtain a generalized synthesis algorithm covering positive and negative causal relations that has also been implemented in the RTA workflow. For the reader familiar with the foundations of Petri nets, it is also worth to point out that structurally causal networks of depth $2$ are a subclass of Petri nets, and it might be possible to establish a formal connection between the generalized Petri nets called piles in \cite{Petri96} and deeper causal networks.

\section{Implementation in JupyterFlow}

Motivated by the need to interact with subject matter experts, the need for reproducibility, and the need to efficiently execute and validate machine learning algorithms on large data sets, we have developed at SRI what is to our knowledge the first integrated approach that combines the benefits of interactive computing and distributed workflows. 
This framework, named \textit{JupyterFlow}, is based on Petri nets (generalized dataflow graphs) \cite{petrinets} and integrates features of interactive computing (Jupyter/Python Notebooks) \cite{jupyter} with data-driven distributed and parallel execution on heterogenous clusters, typically with GPUs (Graphical Processing Units) on a subset of the nodes. JupyterFlow integrates with TensorFlow \cite{tensorflow}, which in the RTA workflow is used for estimating Gaussian Process models \cite{gpflow}, for training a wide range of neural networks, and most recently for approximate
probabilistic inference \cite{palo}.

\begin{figure*}
	\centering 
	\includegraphics[width=1.0\linewidth]{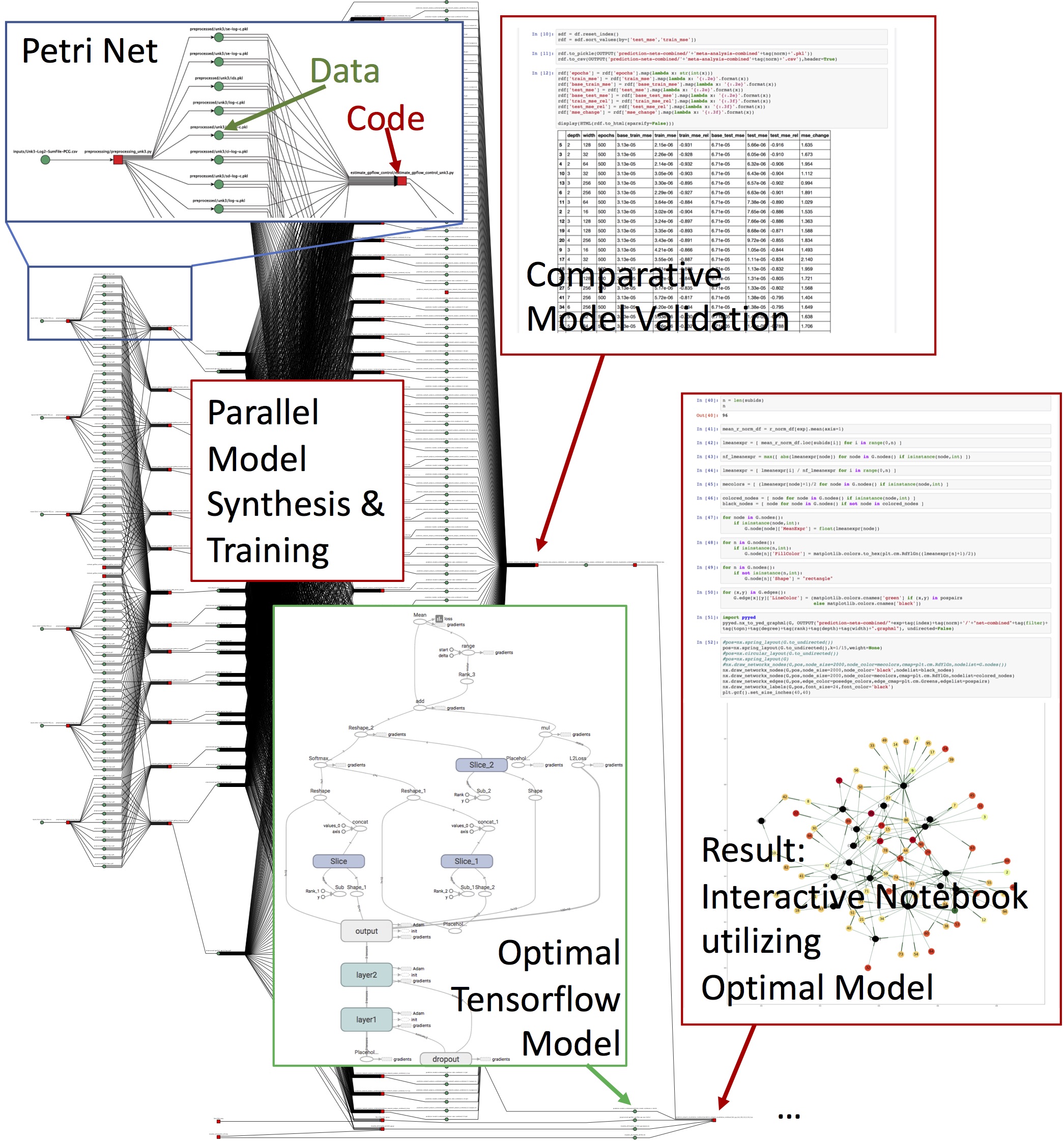}
	\caption{Small Excerpt of the RTA Workflow in JupyterFlow. Each transition in the Petri net is defined by a Python/Jupyter notebook. The excerpt shows part of our graph synthesis algorithm (in fact only the pretraining) utilizing deep and wide neural networks and includes multiple stages: preprocessing, Gaussian process estimation, neural network training, comparative validation, and (intermediate) graph synthesis. The high degree of parallelism admitted by the workflow is quite typical for all our machine learning workflows and important to obtain reasonable execution times. Here is is mainly due to the need for architectural hyperparameter exploration (in the middle), but also the preprocessing and Gaussian process estimation is done in parallel (on the left-hand side).}
	\vspace{-10pt}
	\label{rta-workflow}
\end{figure*}

JupyterFlow supports multiple modes of operations: the parallel execution of workflows on 
powerful GPU servers, the distributed execution on heterogeneous clusters and the cloud (where GPUs may be only available on some machines), and a mode of disconnected operation, where laptops may occasionally synchronize with the main workflow, but otherwise can work independently (without continuous network connectivity). Generally, the workflow executes automatically as a collection of computational notebooks with automatically inferred dataflow dependencies and resource constraints (e.g., memory, number of GPUs), which are essential to optimize the location of computations in the cluster. In addition, each notebook can also be executed interactively (e.g., for diagnosis and experimentation). 

Just as TensorFlow utilizes a form of dataflow graphs in the small, JupyterFlow utilizes dataflow graphs in the large.  In the case of RTA, our current workflow (see Fig. \ref{rta-workflow} for a small excerpt) consists of $\approx 25,000$ notebooks utilizing Python and TensorFlow that give rise to $\approx 500,000$ metalevel dataflow dependencies (this is not counting dependencies at the TensorFlow level). We are currently executing this workflow on SRI's private cluster (with up to 4 GPUs per node) and in the Google Cloud (using large instances with up to 96 CPUs and 8 GPUs). Our latest version of JupyterFlow also runs in Docker containers and supports virtual heterogeneous  Kubernetes clusters (we tested up to 100 nodes) which can be scaled up and down dynamically. Efficiently sharing large amounts of data becomes possible by taking advantage of the combination of an overlay file system and the powerful possibilities offered by shared storage devices in the Google cloud.

\section{Conclusion}

Viewing biological systems as complex continuous systems that can be modeled as networks of interacting components at a very abstract level, we have presented three causal network synthesis algorithms, that support the biologist in
understanding the mechanism of action (MoA) of a system perturbation, which in our experiments were caused by a toxin or a drug. All our algorithms take into account the reality that only a very limited period after the perturbation is relevant to identifying the MoA. Interestingly, this limitation is not unique to perturbations in the biological domain. The lack of data due to the limited time period is partly compensated by the possibility of performing a large number of measurements in parallel with today's high-throughput technologies, leading to a relatively short but high-dimensional time series for each experiment. Nevertheless, our modeling has to be sufficiently abstract to account for the limited amount of data, which in absence of specific biological knowledge precludes detailed stochastic and differential equation models. 

All our algorithms combine Gaussian processes with abstract network modeling. Our probabilistic causal network synthesis algorithm is based on an intermediate synthetic model that tries to capture key aspects of causality with very few parameters. Our autoencoder-based network synthesis does not give approximate probabilistic results but can use generic knowledge about causality to identify models that are more likely (based on rough probabilistic lower bounds) to be consistent with the data. Finally, our deep neural network prediction-based algorithm is purely heuristic, but allows networks that can express more detail about the dynamic nature of interactions.

In this paper we have discussed the first probabilistic synthesis algorithm in more detail. This is because
of its simplicity, the low complexity of the underlying model, and its capability to generate at least approximate probabilities,
which were confirmed to be roughly biologically plausible. We believe that the other two algorithms, while potentially more powerful in the detection of causalities (and hence still interesting for hypothesis generation), may be more prone to false positives, due to their relatively higher complexity, but we think that more experiments (possibly beyond the biological domain) are needed to understand their relative advantages and limitations.

Here, we briefly summarize a number of extensions of our probabilistic synthesis algorithm that might be worthwhile to explore. First, a natural extension is to take into account inhibitor edges (that we also refer to as negative causal edges in contrast to the positives ones). Such negative causal edges are already implemented in our other heuristic algorithms. Based on the Connectivity Map data, we know that such negative dependencies are much rarer, but they can be useful to complement the picture, and also lead to a more precise notion of independence, if defined as the absence of causality.

There are at least two places where with minor modifications our synthetic model can take into account additional biological knowledge about the network. First, by making additional assumptions about the dynamics of causality, time lags could be modeled using a parameterized distribution as in \cite{Ramsey08}. Although, in our view, worthwhile to investigate, the price to pay for this approach is that it will lead to different models (and with varying degree of uncertainty) depending of the type of data, e.g. transcriptomics, proteomics, and metabolomics, complicating an integrated network-based treatment. The other possible improvement is concerned with the balanced nature of the synthetic data set. A prior reflecting a lower probability of causality vs. independence, may slightly improve the accuracy, but in the light of the typical use of the causality detector, e.g., when applied after some filtering to significantly affected pairs of genes, the prior may not be accurate anymore, leading to another potential source of error. Again different types of data would also have to be taken into account, leading to increased complexity of the model. Another interesting avenue for future work is to consider Generative Adversarial Networks (GANs) \cite{gans} as an alternative to sampling based on Gaussian processes. We have already used GANs for anomaly detection in the RTA workflow, but not directly exploited their generative capabilities. It would be interesting to understand if especially the convolutional GANs that we are using can lead to biologically more realistic modeling (especially regarding the overall distribution) and hence potential accuracy improvements for our Siamese networks.

Finally, we would like to mention the opportunity to use our probabilistic causality detector as part of another algorithm, which can lead to further improvements (e.g. reduction of false positive rates). In the probabilistic algorithm presented in this paper, causality detection (both the existence of causality and its direction) is performed locally (and hence without taking into account network context), that is only on the basis of two potentially related time series. Global knowledge about network properties, e.g. relative consistency between edges, absence of cycles, or degree/density knowledge cannot be exploited in this way. A possible systematic solution is an integrated treatment of data and knowledge in machine learning using PALO, our Probabilistic Approximate Logic that is subject of a separate paper \cite{palo}, which also includes the application to biological network synthesis but has broader applicability.

Due to the small number of assumptions underlying our network synthesis algorithms, we expect that they also may be applicable in other domains where a complex system is perturbed and high-dimensional time series data for a relative small number of time points after the perturbation needs to be analyzed to understand the mechanism of action, in other words, how the perturbation propagates through the system. One sample domain, for which we obtained some promising preliminary results, is the domain of global financial markets, where the high dimensionality of the time series results from the broad range of tradable financial products that can be seen as observables of the global financial system. This is 
another quite natural test case for our algorithms, since financial time series data is readily available and there are many types of perturbation events that are reoccurring and well-defined. Previous work indicates that network modeling can have important applications in this area \cite{Lopez14}. Other domains of potential interest are related to social media and news, where the high dimensionality of the observation space is a natural result of the vast number of interrelated topics and information sources. For example, we have done some preliminary semantic analysis and time series modeling using the very rich data source of global events and news is provided by the GDELT project \cite{GDELT}.  Understanding the network-scale effects of events could lead to more accurate assessment of news content, its sources, and their inherent biases. 
The vast amount of available data is reflected in a larger amount of computing and storage resources needed in spite of our use of lightweight natural language processing techniques. A restriction to subsets of the full media universe (e.g. to particular domains of interest) turned out to be more tractable for our initial experiments, which already point to numerous predictable temporal patterns and invariants in the global news cycle.

\vspace*{1ex}

\noindent
{\bf Acknowledgements}
Research was sponsored by the U.S. Army Research Office and the Defense Advanced Research Projects Agency and was accomplished under Cooperative Agreement Number W911NF-14-2-0020. The views and conclusions contained in this document are those of the authors and should not be interpreted as representing the official policies, either expressed or implied, of the Army Research Office, DARPA, or the U.S. Government. The U.S. Government is authorized to reproduce and distribute reprints for Government purposes notwithstanding any copyright notation hereon. We gratefully acknowledge the inspiration and feedback from all members of the RTA project team, especially for their help in interpreting the results and suggesting to explore new directions such as the application of deep neural networks to RTA data sets.

\vspace*{1ex}

\noindent
{\bf Author Contributions}
Akos Vertes, Andrew Poggio, and Carolyn L. Talcott conceived the research. Mark-Oliver Stehr conceived and implemented the JupyterFlow framework, developed the network synthesis algorithms within this framework, and wrote the manuscript with input from Carolyn L. Talcott and Merrill Knapp. To test the algorithms, Ziad J. Sahab, Peter Avar, Andrew R. Korte, Lida Parvin, and Akos Vertes carried out the metabolomics and proteomics analysis. Deborah I. Bunin and Denise Nishita carried out the transcriptomics analysis. Akos Vertes, Ziad J. Sahab, Peter Avar, Andrew R. Korte, Mark-Oliver Stehr, Carolyn L. Talcott, Merrill Knapp, Andrew Poggio, Deborah I. Bunin, Brian M. Davis, Christine A. Morton, Christopher J. Sevinsky, and Maria I. Zavodszky contributed to the discussions of the results.

\bibliographystyle{abbrv}
\bibliography{ref1,ref2,ref3}

\begin{thebibliography}{10}

\bibitem{tensorflow}
M.~Abadi, P.~Barham, J.~Chen, Z.~Chen, A.~Davis, J.~Dean, M.~Devin,
  S.~Ghemawat, G.~Irving, M.~Isard, M.~Kudlur, J.~Levenberg, R.~Monga,
  S.~Moore, D.~G. Murray, B.~Steiner, P.~Tucker, V.~Vasudevan, P.~Warden,
  M.~Wicke, Y.~Yu, and X.~Zheng.
\newblock Tensorflow: A system for large-scale machine learning.
\newblock In {\em Proceedings of the 12th USENIX Conference on Operating
  Systems Design and Implementation}, OSDI'16, pages 265--283, Berkeley, CA,
  USA, 2016. USENIX Association.

\bibitem{Albert05}
R.~Albert.
\newblock Scale-free networks in cell biology.
\newblock {\em Journal of Cell Science}, 118(21):4947--4957, 2005.

\bibitem{Berlemont18}
S.~Berlemont, G.~Lefebvre, S.~Duffner, and C.~Garcia.
\newblock Class-balanced {Siamese} neural networks.
\newblock {\em Neurocomput.}, 273(C):47--56, Jan. 2018.

\bibitem{Bromley93}
J.~Bromley, J.~W. Bentz, L.~Bottou, I.~Guyon, Y.~LeCun, C.~Moore,
  E.~S{\"{a}}ckinger, and R.~Shah.
\newblock Signature verification using {A} "{Siamese}" time delay neural
  network.
\newblock {\em {IJPRAI}}, 7(4):669--688, 1993.

\bibitem{Bromley94}
J.~Bromley, I.~Guyon, Y.~LeCun, E.~S\"{a}ckinger, and R.~Shah.
\newblock Signature verification using a "{Siamese}" time delay neural network.
\newblock In J.~D. Cowan, G.~Tesauro, and J.~Alspector, editors, {\em Advances
  in Neural Information Processing Systems 6}, pages 737--744. Morgan-Kaufmann,
  1994.

\bibitem{Lopez14}
N.~J.~C. Calkin and M.~López~de Prado.
\newblock The topology of macro financial flows: An application of stochastic
  flow diagrams.
\newblock {\em Algorithmic Finance}, 3(1-2):43--85, 2014.

\bibitem{Chaouiya07}
C.~Chaouiya.
\newblock {Petri net modelling of biological networks}.
\newblock {\em Briefings in Bioinformatics}, 8(4):210--219, 07 2007.

\bibitem{Chu08}
T.~Chu and C.~Glymour.
\newblock Search for additive nonlinear time series causal models.
\newblock {\em J. Mach. Learn. Res.}, 9:967--991, June 2008.

\bibitem{gpflow}
A.~G. de~G.~Matthews, M.~van~der Wilk, T.~Nickson, K.~Fujii, A.~Boukouvalas,
  P.~Le{\'{o}}n{-}Villagr{\'{a}}, Z.~Ghahramani, and J.~Hensman.
\newblock {GPflow}: {A} {Gaussian} process library using {TensorFlow}.
\newblock {\em Journal of Machine Learning Research}, 18:40:1--40:6, 2017.

\bibitem{Droghini18}
D.~Droghini, F.~Vesperini, E.~Principi, S.~Squartini, and F.~Piazza.
\newblock Few-shot {Siamese} neural networks employing audio features for
  human-fall detection.
\newblock In {\em Proceedings of the International Conference on Pattern
  Recognition and Artificial Intelligence}, PRAI 2018, pages 63--69, New York,
  NY, USA, 2018. ACM.

\bibitem{GDELT}
{GDELT: Global Database of Events, Language, and Tone}.
\newblock \url{https://www.gdeltproject.org/about.html}.

\bibitem{Gerstein12}
M.~B. Gerstein, A.~Kundaje, M.~Hariharan, S.~G. Landt, K.~K. Yan, C.~Cheng,
  X.~J. Mu, E.~Khurana, J.~Rozowsky, R.~Alexander, R.~Min, P.~Alves, A.~Abyzov,
  N.~Addleman, N.~Bhardwaj, A.~P. Boyle, P.~Cayting, A.~Charos, D.~Z. Chen,
  Y.~Cheng, D.~Clarke, C.~Eastman, G.~Euskirchen, S.~Frietze, Y.~Fu, J.~Gertz,
  F.~Grubert, A.~Harmanci, P.~Jain, M.~Kasowski, P.~Lacroute, J.~J. Leng,
  J.~Lian, H.~Monahan, H.~O'Geen, Z.~Ouyang, E.~C. Partridge, D.~Patacsil,
  F.~Pauli, D.~Raha, L.~Ramirez, T.~E. Reddy, B.~Reed, M.~Shi, T.~Slifer,
  J.~Wang, L.~Wu, X.~Yang, K.~Y. Yip, G.~Zilberman-Schapira, S.~Batzoglou,
  A.~Sidow, P.~J. Farnham, R.~M. Myers, S.~M. Weissman, and M.~Snyder.
\newblock Architecture of the human regulatory network derived from {ENCODE}
  data.
\newblock {\em Nature}, 489(7414):91--100, 2012.

\bibitem{petrinets}
C.~Girault and R.~Valk.
\newblock {\em Petri Nets for Systems Engineering: A Guide to Modeling,
  Verification, and Applications}.
\newblock Springer Publishing Company, Incorporated, 1st edition, 2010.

\bibitem{gans}
I.~J. Goodfellow, J.~Pouget-Abadie, M.~Mirza, B.~Xu, D.~Warde-Farley, S.~Ozair,
  A.~Courville, and Y.~Bengio.
\newblock Generative adversarial nets.
\newblock In {\em Proceedings of the 27th International Conference on Neural
  Information Processing Systems - Volume 2}, NIPS'14, pages 2672--2680,
  Cambridge, MA, USA, 2014. MIT Press.

\bibitem{Gosak18}
M.~{Gosak}, R.~{Markovi{\v c}}, J.~{Dolen{\v s}ek}, M.~{Slak Rupnik},
  M.~{Marhl}, A.~{Sto{\v z}er}, and M.~{Perc}.
\newblock {Network science of biological systems at different scales: A
  review}.
\newblock {\em Physics of Life Reviews}, 24:118--135, Mar. 2018.

\bibitem{curriculum-learning}
A.~Graves, M.~G. Bellemare, J.~Menick, R.~Munos, and K.~Kavukcuoglu.
\newblock Automated curriculum learning for neural networks.
\newblock In {\em Proceedings of the 34th International Conference on Machine
  Learning, {ICML} 2017, Sydney, NSW, Australia, 6-11 August 2017}, pages
  1311--1320, 2017.

\bibitem{Han15}
H.~Han, H.~Shim, D.~Shin, J.~E. Shim, Y.~Ko, J.~Shin, H.~Kim, A.~Cho, E.~Kim,
  T.~Lee, H.~Kim, K.~Kim, S.~Yang, D.~Bae, A.~Yun, S.~Kim, C.~Y. Kim, H.~J.
  Cho, B.~Kang, S.~Shin, and I.~Lee.
\newblock {TRRUST}: a reference database of human transcriptional regulatory
  interactions.
\newblock {\em Scientific Reports}, 5, 2015.

\bibitem{autoencoders}
G.~E. Hinton and R.~R. Salakhutdinov.
\newblock Reducing the dimensionality of data with neural networks.
\newblock {\em Science}, 313(5786):504--507, July 2006.

\bibitem{jupyter}
T.~Kluyver, B.~Ragan-Kelley, F.~P{\'e}rez, B.~Granger, M.~Bussonnier,
  J.~Frederic, K.~Kelley, J.~Hamrick, J.~Grout, S.~Corlay, P.~Ivanov, D.~Avila,
  S.~Abdalla, and C.~Willing.
\newblock Jupyter notebooks -- a publishing format for reproducible
  computational workflows.
\newblock In F.~Loizides and B.~Schmidt, editors, {\em Positioning and Power in
  Academic Publishing: Players, Agents and Agendas}, pages 87 -- 90. IOS Press,
  2016.

\bibitem{Koch15}
G.~Koch, R.~Zemel, and R.~Salakhutdinov.
\newblock Siamese neural networks for one-shot image recognition.
\newblock In {\em ICML 2015 Deep Learning Workshop}, 2015.

\bibitem{Lamb06}
J.~Lamb, E.~Crawford, D.~Peck, J.~Modell, I.~Blat, M.~Wrobel, J.~Lerner,
  J.~Brunet, A.~Subramanian, K.~Ross, M.~Reich, H.~Hieronymus, G.~Wei,
  S.~Armstrong, S.~Haggarty, P.~Clemons, R.~Wei, S.~Carr, E.~Lander, and
  T.~Golub.
\newblock {The Connectivity Map: using gene-expression signatures to connect
  small molecules and genes and disease.}
\newblock {\em Science}, 313(5795):1929--35, 2006.

\bibitem{Li19}
Q.~Li, J.~Zhu, R.~Cao, K.~Sun, J.~M. Garibaldi, Q.~Li, B.~Liu, and G.~Qiu.
\newblock Relative geometry-aware {Siamese} neural network for {6DOF} camera
  relocalization.
\newblock {\em CoRR}, abs/1901.01049, 2019.

\bibitem{Manocha18}
P.~Manocha, R.~Badlani, A.~Kumar, A.~Shah, B.~Elizalde, and B.~Raj.
\newblock Content-based representations of audio using {Siamese} neural
  networks.
\newblock In {\em {ICASSP}}, pages 3136--3140. {IEEE}, 2018.

\bibitem{Marbach12}
D.~Marbach, J.~C. Costello, R.~Küffner, N.~M. Vega, R.~J. Prill, D.~M.
  Camacho, K.~R. Allison, M.~DREAM5~Consortium, Kellis, J.~J. Collins, and
  G.~Stolovitzky.
\newblock Wisdom of crowds for robust gene network inference.
\newblock {\em Nature methods}, 9(8):796--804, 2012.

\bibitem{conv-autoencoders}
J.~Masci, U.~Meier, D.~Cire\c{s}an, and J.~Schmidhuber.
\newblock Stacked convolutional auto-encoders for hierarchical feature
  extraction.
\newblock In {\em Proceedings of the 21th International Conference on
  Artificial Neural Networks - Volume Part I}, ICANN'11, pages 52--59, Berlin,
  Heidelberg, 2011. Springer-Verlag.

\bibitem{Pearl00}
J.~Pearl.
\newblock {\em Causality: Models, Reasoning, and Inference}.
\newblock Cambridge University Press, New York, NY, USA, 2000.

\bibitem{Petri96}
C.~A. Petri.
\newblock Nets, time and space.
\newblock {\em Theor. Comput. Sci.}, 153(1-2):3--48, Jan. 1996.

\bibitem{Prill10}
R.~J. Prill, D.~Marbach, J.~Saez-Rodriguez, P.~K. Sorger, L.~G. Alexopoulos,
  X.~Xue, N.~D. Clarke, G.~Altan-Bonnet, and G.~Stolovitzky.
\newblock Towards a rigorous assessment of systems biology models: the {DREAM3}
  challenges.
\newblock {\em PloS one}, 5(2), 2010.

\bibitem{Ramsey08}
S.~A. Ramsey, S.~L. Klemm, D.~E. Zak, K.~A. Kennedy, V.~Thorsson, B.~Li,
  M.~Gilchrist, E.~S. Gold, C.~D. Johnson, V.~Litvak, G.~Navarro, J.~C. Roach,
  C.~M. Rosenberger, A.~G. Rust, N.~Yudkovsky, A.~Aderem, and I.~Shmulevich.
\newblock Uncovering a macrophage transcriptional program by integrating
  evidence from motif scanning and expression dynamics.
\newblock {\em PLoS Computational Biology}, 4(3), 2008.

\bibitem{gaussianprocesses}
C.~E. Rasmussen and C.~K.~I. Williams.
\newblock {\em Gaussian Processes for Machine Learning (Adaptive Computation
  and Machine Learning)}.
\newblock The MIT Press, 2005.

\bibitem{Shaham15}
U.~Shaham and R.~R. Lederman.
\newblock Common variable learning and invariant representation learning using
  {Siamese} neural networks.
\newblock {\em CoRR}, abs/1512.08806, 2015.

\bibitem{ShahamL18}
U.~Shaham and R.~R. Lederman.
\newblock Learning by coincidence: {Siamese} networks and common variable
  learning.
\newblock {\em Pattern Recognition}, 74:52--63, 2018.

\bibitem{Spirtes93}
P.~Spirtes, C.~Glymour, and R.~Scheines.
\newblock {\em Causation, Prediction, and Search}.
\newblock Springer, 1993.

\bibitem{palo}
M.-O. Stehr.
\newblock {Probabilistic Approximate Logic and its Implementation in the
  Logical Imagination Engine}, 2019.
\newblock {SRI International Technical Report (in preparation)}.

\bibitem{Subramanian17}
A.~Subramanian, R.~Narayan, S.~M. Corsello, D.~D. Peck, T.~E. Natoli, X.~Lu,
  J.~Gould, J.~F. Davis, A.~A. Tubelli, J.~K. Asiedu, D.~L. Lahr, J.~E.
  Hirschman, Z.~Liu, M.~Donahue, B.~Julian, M.~Khan, D.~Wadden, I.~C. Smith,
  D.~Lam, A.~Liberzon, C.~Toder, M.~Bagul, M.~Orzechowski, O.~M. Enache,
  F.~Piccioni, S.~A. Johnson, N.~J. Lyons, A.~H. Berger, A.~F. Shamji, A.~N.
  Brooks, A.~Vrcic, C.~Flynn, J.~Rosains, D.~Y. Takeda, R.~Hu, D.~Davison,
  J.~Lamb, K.~Ardlie, L.~Hogstrom, P.~Greenside, N.~S. Gray, P.~A. Clemons,
  S.~Silver, X.~Wu, W.~N. Zhao, W.~Read-Button, X.~Wu, S.~J. Haggarty, L.~V.
  Ronco, J.~S. Boehm, S.~L. Schreiber, J.~G. Doench, J.~A. Bittker, D.~E. Root,
  B.~Wong, and T.~R. Golub.
\newblock {A Next Generation Connectivity Map: L1000 Platform and the First
  1,000,000 Profiles}.
\newblock {\em Cell}, 171(6):1437--1452, 2017.

\bibitem{Vertes18}
A.~Vertes, A.~Arul, P.~Avar, A.~R. Korte, H.~Li, P.~Nemes, L.~Parvin,
  S.~Stopka, S.~Hwang, Z.~J. Sahab, L.~Zhang, D.~I. Bunin, M.~Knapp, A.~Poggio,
  M.-O. Stehr, C.~L. Talcott, B.~M. Davis, S.~R. Dinn, C.~A. Morton, C.~J.
  Sevinsky, and M.~I. Zavodszky.
\newblock Inferring mechanism of action of an unknown compound from time series
  omics data.
\newblock In M.~Ceska and D.~Safr{\'{a}}nek, editors, {\em Computational
  Methods in Systems Biology - 16th International Conference, {CMSB} 2018,
  Brno, Czech Republic, September 12-14, 2018, Proceedings}, volume 11095 of
  {\em Lecture Notes in Computer Science}, pages 238--255. Springer, 2018.

\bibitem{Zak09}
D.~E. Zak and A.~Aderem.
\newblock Systems biology of innate immunity.
\newblock {\em Immunological reviews}, 227(1):264--282, 2009.

\bibitem{Zhang18}
Y.~Zhang and Z.~Duan.
\newblock Iminet: Convolutional semi-siamese networks for sound search by vocal
  imitation.
\newblock In {\em {IEEE Workshop on Applications of Signal Processing to Audio
  and Acoustics (WASPAA)}}, pages 304--308. {IEEE}, 2018.

\end{thebibliography}

\end{document}